\documentclass[11pt, a4paper, logo, copyright, nonumbering]{deepseek}
\usepackage[authoryear, sort&compress, round]{natbib}
\usepackage{dblfloatfix}
\usepackage{ulem}
\usepackage{caption}
\usepackage{dramatist}
\usepackage{xspace}
\usepackage{pifont}
\usepackage{multirow}
\usepackage{tcolorbox}
\usepackage{xltabular}
\usepackage{longtable}
\usepackage{pdfpages}
\usepackage{xcolor}
\definecolor{linkcolor}{rgb}{0.956,0.298,0.235}
\definecolor{citecolor}{HTML}{1976D2}
\usepackage{tabularx}
\usepackage{hyperref}
\hypersetup{
    colorlinks=true,    
    linkcolor=citecolor,     
    filecolor=magenta,  
    urlcolor=cyan,      
    citecolor=citecolor,    
}

\usepackage{amsfonts}
\usepackage{amsmath}
\usepackage{amssymb}
\usepackage{lineno}

\usepackage[bottom]{footmisc}

\usepackage{CJKutf8}
\usepackage{subfigure}
\usepackage{setspace}

\newcommand{\eg}{e.g.,\xspace}

\usepackage{xspace}

\makeatletter
\def\@BTrule[#1]{%
  \ifx\longtable\undefined
    \let\@BTswitch\@BTnormal
  \else\ifx\hline\LT@hline
    \nobreak
    \let\@BTswitch\@BLTrule
  \else
     \let\@BTswitch\@BTnormal
  \fi\fi
  \global\@thisrulewidth=#1\relax
  \ifnum\@thisruleclass=\tw@\vskip\@aboverulesep\else
  \ifnum\@lastruleclass=\z@\vskip\@aboverulesep\else
  \ifnum\@lastruleclass=\@ne\vskip\doublerulesep\fi\fi\fi
  \@BTswitch}
\makeatother

\addto\extrasenglish{
}

 {\begin{list}{}%
         {\setlength{\leftmargin}{#1}}%
         \item[]%
 }
 {\end{list}}
 
\reportnumber{001} 

\renewcommand{\today}{}

\title{\centering DeepSeek-VL: Towards Real-World Vision-Language Understanding}

\author[*]{
\small

Haoyu Lu*$^{1\dag}$,
Wen Liu*$^{1}$,
Bo Zhang*$^{1\ddag}$,
Bingxuan Wang$^{1\dag}$,
Kai Dong$^{1}$, 
Bo Liu$^{1\dag}$,
Jingxiang Sun$^{1\dag}$,
Tongzheng Ren$^{1\dag}$,
Zhuoshu Li$^{1}$,
Hao Yang$^{1\dag}$,  
Yaofeng Sun$^{1}$,
Chengqi Deng$^{1}$,
Hanwei Xu$^{1}$,
Zhenda Xie$^{1}$,
Chong Ruan$^{1}$

\small
$^1$DeepSeek-AI \\
\small
\texttt{\{neal, liuwen, bo\}@deepseek.com} \\
\small
\url{https://github.com/deepseek-ai/DeepSeek-VL}
}
\correspondingauthor{\footnotesize\, $^*$ Equal contribution. \\\, $^\dag$ Work done during the internship at DeepSeek-AI. \\\, $^\ddag$ Project lead.}

\renewcommand{\phi}{\varphi}

\renewcommand{\epsilon}{\varepsilon}
\renewcommand{\imath}{\mathrm{i}}

\newlength{\restsubwidth}
\newlength{\restsubheight}
\newlength{\restsubmoreheight}
\newcommand{\rest}[2]{%
        \settowidth{\restsubwidth}{\ensuremath{#2}}
        \settoheight{\restsubheight}{\ensuremath{{}_{#2}}}
        \ensuremath{{#1\hskip 0.5pt}_{\vrule\kern2pt\parbox[b][%
        4pt][b]{\the\restsubwidth}{%
                        \ensuremath{{}_{#2}}}}}
        }

\begin{abstract}

We present DeepSeek-VL, an open-source Vision-Language (VL) Model designed for real-world vision and language understanding applications. Our approach is structured around three key dimensions:

\noindent\hspace{1em}\textbullet\hspace{0.5em}\textbf{Data Construction}: We strive to ensure our data is diverse, scalable and extensively covers real-world scenarios including web screenshots, PDFs, OCR, charts, and knowledge-based content (expert knowledge, textbooks), aiming for a comprehensive representation of practical contexts. Further, we create a use case taxonomy from real user scenarios and construct an instruction-tuning dataset accordingly. The fine-tuning with this dataset substantially improves the model's user experience in practical applications.

\noindent\hspace{1em}\textbullet\hspace{0.5em}\textbf{Model Architecture}: Considering efficiency and the demands of most real-world scenarios, DeepSeek-VL incorporates a hybrid vision encoder that efficiently processes high-resolution images (1024 x 1024) within a fixed token budget, while maintaining a relatively low computational overhead. This design choice ensures the model's ability to capture critical semantic and detailed information across various visual tasks.

\noindent\hspace{1em}\textbullet\hspace{0.5em}\textbf{Training Strategy}: We posit that a proficient Vision-Language Model should, foremost, possess strong language abilities. To ensure the preservation of LLM capabilities during pretraining, we investigate an effective VL pretraining strategy by integrating LLM training from the beginning and carefully managing the competitive dynamics observed between vision and language modalities. Starting with a focus on text, we gradually adjust the ratio to facilitate a balanced integration of both modalities.

The DeepSeek-VL family (both 1.3B and 7B models) showcases superior user experiences as a vision-language chatbot in real-world applications, achieving state-of-the-art or competitive performance across a wide range of visual-language benchmarks at the same model size while maintaining robust performance on language-centric benchmarks. We have made both 1.3B and 7B models publicly accessible to foster innovations based on this foundation model.

\end{abstract}

\begin{document}
\begin{CJK*}{UTF8}{gbsn}

\maketitle

\newpage

\begin{spacing}{0.9}
\tableofcontents
\end{spacing}

\newpage

\section{Introduction}

The remarkable success of large language models (LLMs)~\citep{chatgpt,gpt4,claude,bard} has fueled the demand for a versatile interface that can handle multiple modalities beyond language. 
In response to this growing demand, we have seen an emergence of Large Multimodal Models (LMMs) like GPT-4V~\citep{gpt4v} and Gemini~\citep{gemini}, which serve as versatile assistants capable of comprehending and acting upon instructions that span vision and language. These models exhibit considerable promise in executing complex, diverse real-world tasks, enabling more natural and human-like interactions.
%


\begin{figure}[p]
    \vspace*{-2cm}
    \makebox[\linewidth]{
        \includegraphics[width=1.22\linewidth]{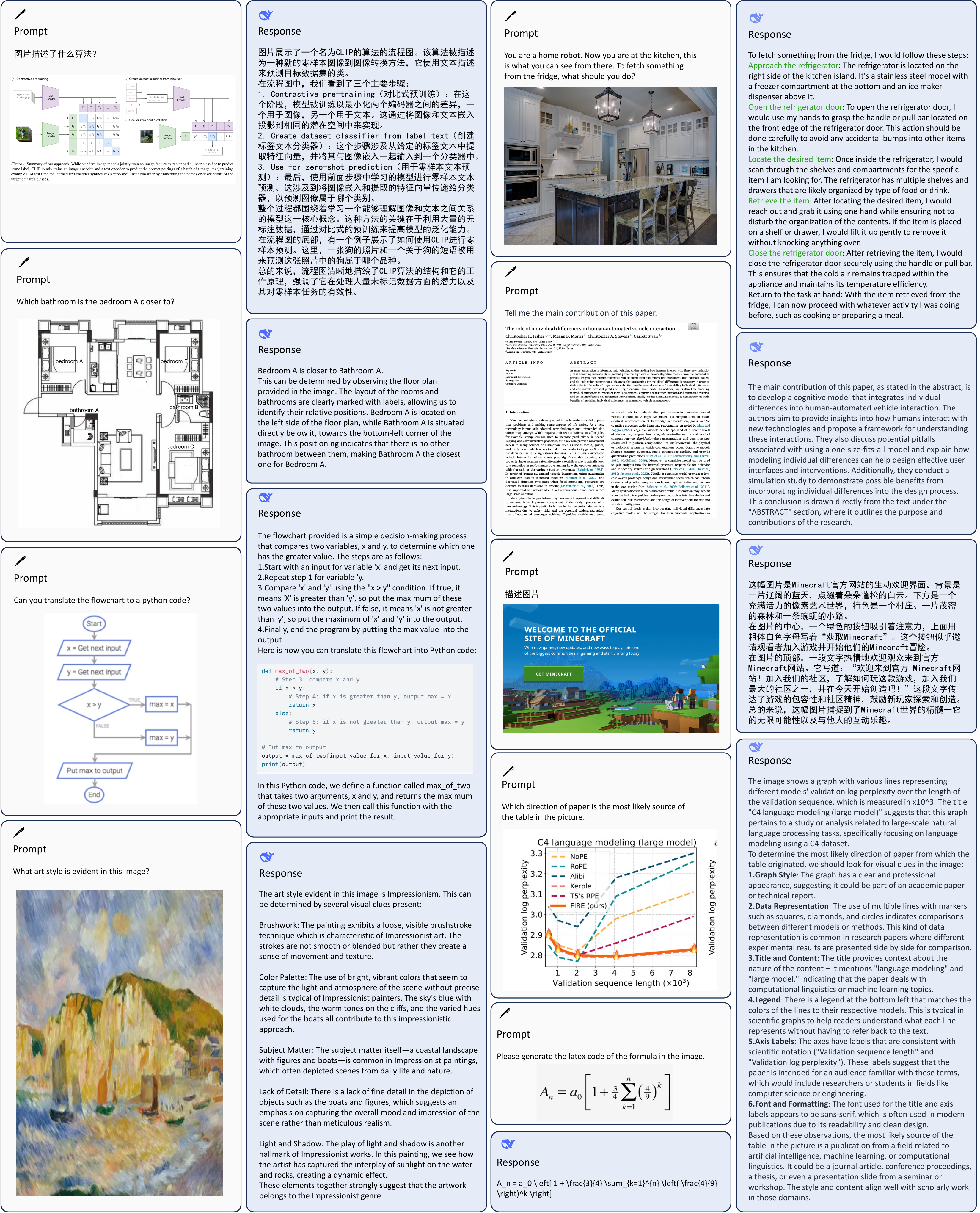}
    }
    \caption{DeepSeek-VL possesses general multimodal understanding capabilities, capable of processing logical diagrams, web pages, formula recognition, scientific literature, natural images, and embodied intelligence in complex scenarios.}
\end{figure}

Recently, there has been a surge of open-source large multimodal models aimed at narrowing the gap with proprietary counterparts. Substantial strides have been made, especially in benchmark performance, yet a significant divide persists between the majority of open-source models and state-of-the-art closed-source models~\citep{gpt4v,fuyu-8b,gemini,qwen-vl} when it comes to real-world performance and user experience. It remains challenging for the open-source community to develop models with robust general multimodal capabilities for real-world applications.

The performance gap between the most open-source models and the proprietary models is largely pronounced in real-world scenarios, primarily due to the following reasons:
\begin{itemize} \itemsep0.5em

\item Many open-source solutions allocate a significant proportion of computational resources to the instruction tuning phase. However, the experience of training powerful language models underscores the importance of extensive pretraining in the development of general intelligence. To imbue multimodal models with rich world knowledge, there should be an emphasis on comprehensive pretraining that leverages a broad spectrum of vision-language data.

\item A common practice is to amalgamate various academic datasets during instruction tuning. While such an approach may yield good benchmark results, it often falls short in providing an authentic real-world usage experience. 

\item In terms of model architecture, prior works mostly adapt a vision transformer, typically text-aligned, to a pre-trained language model. However, most of these models operate on a relatively low resolution, \eg 336$\times$336 or 448$\times$ 448. The intricacies of complex real-world scenarios, such as optical character recognition or tiny object discernment, demand high-resolution processing capability.  

\item 
While some models~\citep{emu,cogvlm,yi-vl-34b,lin2023vila} have begun to exploit pretraining, they often overlook the preservation of language skills. Often, there is a degradation of language capability after prolonged multimodal training. Since we aim for a generalist that possesses strong capabilities in both modalities, there should be a training strategy that well preserves the language capability when developing the new modality ability.

\end{itemize} 

In light of these, we present DeepSeek-VL, an open-source large multimodal model, which is built upon the DeepSeek language model series. We develop the model in the pursuit of adept performance in real-world scenarios, which involves extensive pretraining, careful data curation based on a use case taxonomy, model architecture design for high-resolution processing, and a training strategy that balances the multi-modalities.
On top of these, we develop a training methodology that steers the model scaling, from 1B to 7B. These comprehensive explorations bring a significant performance advantage in practical settings, compared to other large multimodal models (LMMs) of similar size. 

DeepSeek-VL's pretraining dataset is compiled 
from a variety of sources, including but not limited to Common Crawl, Web Code, E-books, Educational Materials, and arXiv Articles. This collection thoroughly encompasses real-world scenarios such as web screenshots, PDFs, OCR, charts, and knowledge-based content (expertise, textbooks), aiming for a broad and practical representation while remaining scalable.

While our pretraining data encompasses a wide array of world knowledge, we meticulously curate our instruction-tuning dataset to reflect real-world usage scenarios. To achieve this, we manually gather authentic test cases for GPT-4V and Gemini from the Internet. These cases have been systematically organized into a comprehensive taxonomy. We use this structured taxonomy to choose prompts for each test image, ensuring a practical and relevant instruction tuning dataset. This taxonomy is also used to create an evaluation dataset that effectively assesses real-world performance.

The visual module is designed to optimize the utilization of high-resolution visual inputs while remaining within a fixed token budget to manage inference costs effectively. As such, we employ a hybrid vision encoder, which combines a text-aligned encoder for coarse semantic extraction at $384\times384$ resolution with a high-resolution encoder that captures detailed visual information at $1024\times1024$ resolution. By fusing these two encoders, our hybrid approach efficiently condenses a 1024×1024 resolution image (which suffices in most use cases) into 576 tokens. This token count strikes a balance between rich visual representation and token economy, making it feasible for both text-image interleaving and multi-turn inference scenarios.

During the pretraining of multimodal models, a common challenge encountered is the potential degradation of language capabilities when the training process is overly reliant on vision-language data. Our research reveals that maintaining a significant proportion of language data—specifically, at least 70\%—is essential to preserve the integrity of language knowledge within the model. This balance is critical for achieving a robust multimodal capability that does not compromise language performance. Moreover, we introduce a novel ``modality warm-up'' strategy. This approach carefully adjusts the ratio of modalities during training, gradually incorporating more vision-language data. 
The careful tuning of the modality ratio along with the warm-up strategy results in a balanced performance of both modalities. 

When iterating on our model, We conduct experiments on a small scale before scaling to a larger model size. However, a smaller model, \eg 1B model, cannot demonstrate reasonable performance on benchmarks~\citep{schaeffer2024emergent} and faithfully reflect the model's performance. We adopt two approaches to address this. First, we modify the evaluation protocol from multi-choice to compare the perplexity of options. Also, to prevent the instruction following ability from becoming the bottleneck, we mix a small proportion of instruction tuning data during the pretraining phase. In this way, we can achieve reasonable performance using the 1B model and more accurately measure the impact of each iteration during the experiment.

Through extensive evaluations of general vision and language benchmarks, the DeepSeek-VL family showcases superior user experiences in real-world applications and achieves state-of-the-art or competitive performance across a wide range of visual-language benchmarks at the same model size, while maintaining robust language-centric performance. To foster innovation and enable a wide range of applications, we have made two versions of our ours, 1.3B and 7B, publicly accessible, in the hope of facilitating the needs of varying computational capabilities.

\section{Data Construction}
A diverse and large dataset is the most important ingredient of visual language model training. Our dataset can be divided into two parts: Vision-Language pretraining Data and Vision-Language Supervised Fine-Tuning Data. VL pretraining Data is composed of visual-text data from various sources, aimed at enhancing the model's fundamental cross-modal understanding capabilities; while VL Supervised Fine-Tuning Data has a relatively smaller size and aims to teach the model to complete specific downstream tasks. By design, VL pretraining Data is used to warm up the vision-language adaptor in training stage 1 and jointly pretrain the vision-language model in stage 2, and VL Supervised Fine-Tuning Data is exploited in training stage 3, i.e., vision language supervised fine-tuning.

\subsection{Vision-Language pretraining Data}

\renewcommand{\arraystretch}{1.1}
\begin{table}[t!]
\centering
\small
\caption{Summary of datasets used in the joint vision and language pretraining stage.}
\label{tab:pretraining_dataset}
\begin{tabular}{llr}
\toprule
\textbf{Category} & \textbf{Dataset} & \textbf{Ratio} \\ 
\midrule
Interleaved image-text & MMC4~\citep{mmc4} & 13.1\% \\ 
 & Wikipedia EN\& CN~\citep{wikidump} &\\ 
 & Wikihow~\citep{wikihow} &\\ 
 & in-house PDF and Epub textbooks &  \\ 
\midrule
Image caption & Capsfusion~\citep{capsfus} & 11.1\%\\ 
 & TaiSu~\citep{liu2022taisu} &  \\ 
 & Detailed Caption~\citep{detailed_caption} &  \\ 
\midrule
Table and chart & Chart2text~\citep{chart2text} & 2.1\%\\ 
 & Geo170K~\citep{geo170k} & \\ 
 & Ureader~\citep{ureader} & \\ 
 & Unichart~\citep{unichart} & \\ 
 & M-paper~\citep{m-paper} & \\ 
 & ScienceQA~\citep{scienceqa} & \\ 
 & ScreenQA~\citep{screenqa} & \\ 
 & SciGraphQA-295K~\citep{scigraphqa-295k} & \\ 
 & Paper2figure100k~\citep{paper2figure100k} & \\ 
 & Widget Captioning~\citep{widget-captioning} & \\ 
 & Screen2words~\citep{screen2words} & \\ 
 & Refexp~\citep{refexp} & \\ 
\midrule
Web Code & Websight~\citep{websight} & 0.4\%\\ 
 & python plots scraped from GitHub notebook &  \\ 
\midrule
Scene text OCR & ArT~\citep{chng2019icdar2019} & 1.2\% \\ 
 & MLT-17~\citep{nayef2017icdar2017} & \\ 
 & LSVT~\citep{sun2019icdar}& \\ 
 & UberText~\citep{UberText} & \\ 
 & Coco-text~\citep{veit2016coco} & \\ 
 & RCTW-17~\citep{shi2017icdar2017} & \\ 
 & ReCTS~\citep{zhang2019icdar} & \\ 
 & TextOCR~\citep{singh2021textocr} & \\ 
 & OpenVINO~\citep{krylov2021open} & \\
 & HierText~\citep{long2022towards} & \\ 
\midrule
Document OCR & arXiv rendered markdown~\citep{nougat}& 2.1\% \\
\midrule
Text-only corpus & DeepSeek-LLM 2T text copus~\citep{deepseek-llm} & 70.0\% \\ 
\bottomrule
\end{tabular}
\end{table}

The pretraining dataset utilized in our study encompasses a diverse range of publicly accessible sources, in addition to a selection of proprietary data. We provide a comprehensive overview of the data sources employed during the joint vision and language pretraining stage in Table~\ref{tab:pretraining_dataset}. Such a dataset can facilitate LLM's comprehension of the entities portrayed in the images.

Furthermore, we present a detailed breakdown of the complete dataset, which is organized into the following categories:

\textbf{Interleaved image-text} data enable the models to have a better capability for in-context learning of multi-modality inputs, and we utilize three public datasets MMC4~\citep{mmc4}, Wiki~\citep{burns2023wiki}, Wikihow~\citep{wikihow} and Epub textbooks.

\textbf{Image caption} data come from three high-quality image-text paired datasets: Capsfusion~\citep{capsfus}, TaiSu~\citep{liu2022taisu} and Detailed Caption~\citep{detailed_caption}.

\textbf{Table and chart} data enable the models to learn the capability for general table and chart image understanding. It encompasses a diverse range of public data sources, including Chart2text~\citep{chart2text}, Geo170K~\citep{geo170k},
Unichart~\citep{unichart}, Ureader~\citep{ureader}, 
M-paper~\citep{m-paper}, ScienceQA~\citep{scienceqa}, ScreenQA~\citep{screenqa}, SciGraphQA-295K~\citep{scigraphqa-295k}, Paper2figure100k~\citep{paper2figure100k}, Widget Captioning~\citep{widget-captioning}, Screen2words~\citep{screen2words}, and Refexp~\citep{refexp}.

\textbf{Web Code} data empowers models with the capability to reconstruct code from graphical interfaces or visual plots. Leveraging Websight~\citep{websight} for UI Inverse Rendering, we adopted a strategy akin to that used in MATCHA~\citep{liu2022matcha} for visual plots inverse rendering. This involved the processing of approximately 1.46 million Jupyter notebooks from the Stack dataset~\citep{DenisKocetkov2023}. By extracting these notebooks and collating all diagrams along with their corresponding preceding code segments, we succeeded in curating a collection featuring 2 million pairs of images and codes. For better data quality, we filter 1.1 million instances, each comprising a singular image coupled with a minimum of 5 lines of code, to constitute our primary training dataset.

\textbf{Document Optical Character Recognition (OCR)} data facilitates the recognition of optical characters at the document level, even in challenging real-world scenarios. To the best of our knowledge, there is currently no publicly available large-scale dataset encompassing both English and Chinese documents. Despite the existence of the publicly accessible small-scale dataset Latex-OCR~\citep{latex-ocr}, we additionally constructed a comprehensive English and Chinese document OCR dataset. It is comprised of two parts: 1): \textbf{arXiv Articles:} We collected source code and compiled PDFs from 1.4 million arXiv articles. Utilizing pre-processing tools from Nougat~\citep{nougat}, we rendered these articles into paired images and texts; 2): 
\textbf{E-books and Educational Materials:} We cleaned 860K English and 180K Chinese e-books from Anna's Archive~\citep{annas-archive} alongside millions of K-12 education exam questions. Subsequently, we employed HTML rendering tools~\citep{wkhtmltopdf} to convert these HTML files with different templates into paired image and text formats.


\textbf{Scene text OCR} data augment the capability of the model to recognize and extract text from images in which the text is integrated into the environment. The dataset is composed of multiple public datasets, including ArT~\citep{chng2019icdar2019}, MLT-17~\citep{nayef2017icdar2017}, LSVT~\citep{sun2019icdar}, UberText~\citep{UberText}, Coco-text~\citep{veit2016coco}, RCTW-17~\citep{shi2017icdar2017}, ReCTS~\citep{zhang2019icdar}, TextOCR~\citep{singh2021textocr}, OpenVINO~\citep{krylov2021open} and HierText~\citep{long2022towards}.

\textbf{Text-only corpus} serves to maintain proficiency in language-centric tasks. In this study, we employ the same text corpus with DeepSeek-LLM~\citep{deepseek-llm}.

\renewcommand{\arraystretch}{1.1}
\begin{table}[t!]
\centering
\small
\caption{Summary of data used in our joint vision and language supervised fine-tuning stage.}
\label{tab:sft_datasets}
\begin{tabular}{llc}
\toprule
\textbf{Class} & \textbf{Dataset} & \textbf{Ratio} \\
\midrule

In-house Data & SFT data based on taxonomy (Figure~\ref{tab:image_understanding_taxonomy}) & 
 10.5\%\\
\midrule
General Multi-modality 
& ShareGPT4V~\citep{sharegpt4v} & 35.5\% \\
 & LAION-GPTV~\citep{laion-gpt4v} & \\
 & LVIS-Instruct4V~\citep{lvis-instruct4v} & \\
 & textOCR-GPT4V~\citep{textocr-gpt4v} & \\
 & LLaVA1.6-GPT4V~\citep{llava-v1-6} & \\
 & IconQA~\citep{lu2021iconqa} \\
\midrule
Table and chart & Ureader~\citep{ureader} & 4.1\%\\

 & Geo170K~\citep{geo170k} & \\
 & ScienceQA~\citep{scienceqa} & \\
\midrule
Web Code & Screen-to-code~\citep{screen-to-code} & 2.0\%\\
 & ScreenQA~\citep{screenqa} & \\
\midrule
Text-only SFT & DeepSeek-LLM~\citep{deepseek-llm} & 47.9\%\\
\bottomrule
\end{tabular}
\end{table}

\begin{table*}
    \tiny
    \centering
    \renewcommand\arraystretch{1.5}
    \setlength{\tabcolsep}{2.5pt}
    \begin{tabularx}{\textwidth}{@{}p{2cm}X@{\hspace{10pt}}p{3cm}X@{}}
    \toprule
    Main Category & Description & Secondary Category & Tertiary Category \\ \midrule
    Recognition & This part of the use cases mainly examines the understanding and description ability of large models for image content, which does not require high knowledge reserve and reasoning ability of the model, and some tasks can be completed using traditional machine learning models. & Global Description & Theme Description, Event/Behavior Description, Location/Scene Description, Emotion/Mood Description, Style Recognition, Food Recognition, Others \\ \cmidrule{3-4}
     &  & Local Description & Pointing Description, Position Description, Person Recognition, Object Attribute Description, Logo Recognition, Counting, Currency Recognition \\ \cmidrule{3-4}
     &  & OCR and Transcription & Printed Text Transcription, Handwritten Text Transcription, Specified Format Transcription, Specified Language Transcription \\ \midrule
    Conversion & This type of use case requires the model to be able to describe and recognize image content, and use specific knowledge (e.g., code knowledge, prompt engineering knowledge) to convert image content into another form. & Image to Code & UI to Code, Chart to Code, Photo to SVG/p64 Encoding, Formula to Code, Flowchart to Code \\ \cmidrule{3-4}
     &  & Image to Text & Image to Prompt, Text Summary, Image-based Creation, Text Interpretation \\ \midrule
    Analysis & This type of use case requires the model to use specific knowledge and logical ability to make reasonable analysis and understanding based on image content, and describe the image according to instructions. & Data Chart Analysis & Graph Interpretation, Table Interpretation \\ \cmidrule{3-4}
     &  & Professional Chart Analysis & Circuit Diagram, Flowchart, Map, Music Score, Financial Chart, Floor Plan, Others \\ \cmidrule{3-4}
     &  & Professional Image Analysis & Sensor Image, Biological and Medical Image, Voiceprint Image, Point Cloud Image \\ \cmidrule{3-4}
     &  & Encyclopedia Knowledge Analysis & Art and Culture Knowledge, Natural Environment Knowledge, Food/Clothing/Housing/Transportation Related Knowledge, Entertainment Related Knowledge, Historical Knowledge \\ \midrule
    Commonsense Reasoning & This type of use case mainly tests the model's understanding and mastery of common sense in life, which requires reasoning based on the interpretation and analysis of image content combined with common sense. & Relationship Reasoning & Interpersonal Relationship, Spatial Relationship, Size Relationship, Species Relationship \\ \cmidrule{3-4}
     &  & Function Reasoning & Hardware Function Reasoning, Software Function Reasoning \\ \cmidrule{3-4}
     &  & Environment Reasoning & Environment State Analysis, Environment-based Behavior Reasoning, Embodied Intelligence \\ \cmidrule{3-4}
     &  & Anomaly Reasoning & Identifying Anomalies in Images, Defect Detection, Accident Judgment \\ \cmidrule{3-4}
     &  & Humor Reasoning & - \\ \cmidrule{3-4}
     &  & Other Commonsense Reasoning & State Reasoning, Cause Reasoning, Attribute Comparison, Optical Illusion, Fun Games, Intention Interpretation, Behavior Prediction \\ \midrule
    Logical Reasoning & This type of use case requires the model to combine the understanding of images, comprehensively use domain knowledge and logical reasoning ability to complete corresponding tasks. & Mathematical Reasoning & Algebra and Operation, Plane Geometry, Solid Geometry \\ \cmidrule{3-4}
     &  & Other Logical Reasoning & Physics, Chemistry, Biology, Code, IQ Questions \\ \midrule
    Evaluation & This type of use case requires the model to evaluate the image content according to specific criteria. & - & Reality Evaluation, Similarity Evaluation, Aesthetic Evaluation, Open-ended Evaluation, Improvement Suggestions \\ \midrule
    Multi-graph & This type of use case examines the model's ability to analyze and understand multiple images. & Temporal Sequence Understanding & Event Prediction, Image Sequencing, Behavior Analysis \\ \cmidrule{3-4}
     &  & Multi-graph Comparison & Attribute Comparison, Image-Text Matching, Finding Associations, Spotting Differences, Image Discrimination \\ \midrule
    Safety & This type of use case examines the model's performance in terms of safety. & - & Suggestive Questioning, Counterfactual Questioning, Prompt Injection \\ \bottomrule
    \end{tabularx}
    \vspace{-2mm}
    \caption{Our taxonomy for the in-house SFT data. The categories covered by our high-quality in-house multi-modality SFT data are comprehensively represented in this taxonomy.}
    \label{tab:image_understanding_taxonomy}
    \vspace{-5mm}
\end{table*}
\subsection{Supervised Fine-tuning Data}
The supervised fine-tuning datasets utilized in our study encompass a diverse range of multi-modality and language data sources, including well-known open-source shared gpt4v datasets such as ShareGPT4V~\citep{sharegpt4v}, LAION-GPTV~\citep{laion-gpt4v}, LVIS-Instruct4V~\citep{lvis-instruct4v}, textOCR-GPT4V~\citep{textocr-gpt4v}, LLaVA1.6-GPT4V~\citep{llava-v1-6} and IconQA~\citep{lu2021iconqa}. Additionally, we incorporate partial table and chart data extracted from pretraining datasets such as Ureader~\citep{ureader}, ScreenQA~\citep{screenqa}, Geo170K~\citep{geo170k}, and ScienceQA~\citep{scienceqa}. Moreover, we integrate the UI Code dataset obtained from Screen-to-code~\citep{screen-to-code} tasks. To enhance the quality of our multi-modality SFT data, we have also curated a portion of high-quality in-house multi-modality SFT data, some of which are in the Chinese language. Our in-house instruction-tuning dataset is meticulously designed to reflect real-world usage scenarios and cover a wide range of tasks. We start by collecting a diverse set of authentic test cases for GPT-4V and Gemini from various online sources. These test cases are then carefully analyzed and organized into a comprehensive taxonomy, which encompasses multiple categories, such as recognition, conversion, analysis, reasoning, evaluation, and safety, as detailed in Table~\ref{tab:image_understanding_taxonomy}. This structured taxonomy serves as a guideline for selecting representative prompts for each test image, ensuring that our instruction-tuning dataset is both practical and relevant to real-world applications. Moreover, this taxonomy is also employed to construct a balanced and comprehensive evaluation dataset, which allows us to effectively assess the model's performance across different tasks and categories. By following this systematic approach, we ensure that the categories covered by our in-house multi-modality SFT data are well-aligned with the taxonomy and representative of real-world usage scenarios. Furthermore, we include the text-only SFT data employed in DeepSeek-LLM~\citep{deepseek-llm} as part of our joint vision and language SFT data.


\section{Approach}


\subsection{Architecture}

Our system contains three modules: a hybrid vision encoder, a vision adaptor, and a language model. We introduce each part in this section.

\textbf{Hybrid Vision Encoder.} 
We employ SigLIP as the vision encoder to extract high-level semantic feature representations from visual inputs. However, we observe that a single SigLIP encoder struggles to address all real-world questions comprehensively. Vision encoders in the CLIP family, including SigLIP, are primarily designed for semantic visual representations but are challenged by ambiguous encoding, resulting in visually distinct images being encoded as similar due to what is referred to as "CLIP-blind pairs"~\cite{concat-dino}. Meanwhile, the CLIP family of models is limited by its relatively low-resolution inputs (e.g., 224 x 224, 336 x 336, 384 x 384, 512 x 512), which hinders their ability to handle tasks requiring more detailed low-level features like dense OCR and visual grounding task.

To address these limitations,
recent researches~\citep{vary, concat-dino, lin2023sphinx} have advocated for the integration of additional vision-only self-supervised encoders, to enhance the visual grounding capabilities of multi-modality models. Building upon previous motivations, 
we additionally utilize a vision-only encoder based on the SAM-B~\citep{sam}, a pre-trained ViTDet~\citep{vitdet} image encoder to process low-level features, which accepts high-resolution 1024 x 1024 image inputs. In addition to the SAM-B encoder, we retain the SigLIP-L vision encoder with low-resolution 384 x 384 image inputs. Consequently, our hybrid vision encoder combines the SAM-B and SigLIP-L encoders, efficiently encoding high-resolution 1024 x 1024 images while preserving both semantic and detailed information. Specifically, a high-resolution SAM-B vision encoder first resizes the image into 1024 x 1024 and results in a 64 x 64 x 256 feature map. 

In the case of a high-resolution feature map of size, 64 x 64 x 256 generated by SAM-B, the VL Adaptor initially interpolates it into a size of 96 x 96 x 256. Subsequently, it employs two convolutional layers with a stride of 2, producing a feature map of 24 x 24 x 1024, and reshapes it to 576 x 1024. Alongside this, the low-resolution feature map of size 576 x 1024 generated by SigLIP-L is concatenated with the high-resolution features, resulting in 576 visual tokens with 2048 dimensions. These visual tokens possess a substantial capacity for enhancing high-level semantic visual recognition and low-level visual grounding tasks. Then they undergo GeLU activation and are directed through an embedding layer to establish a connection with the language model.

\begin{figure}[t!]
\centering
\includegraphics[width=1.0\textwidth]{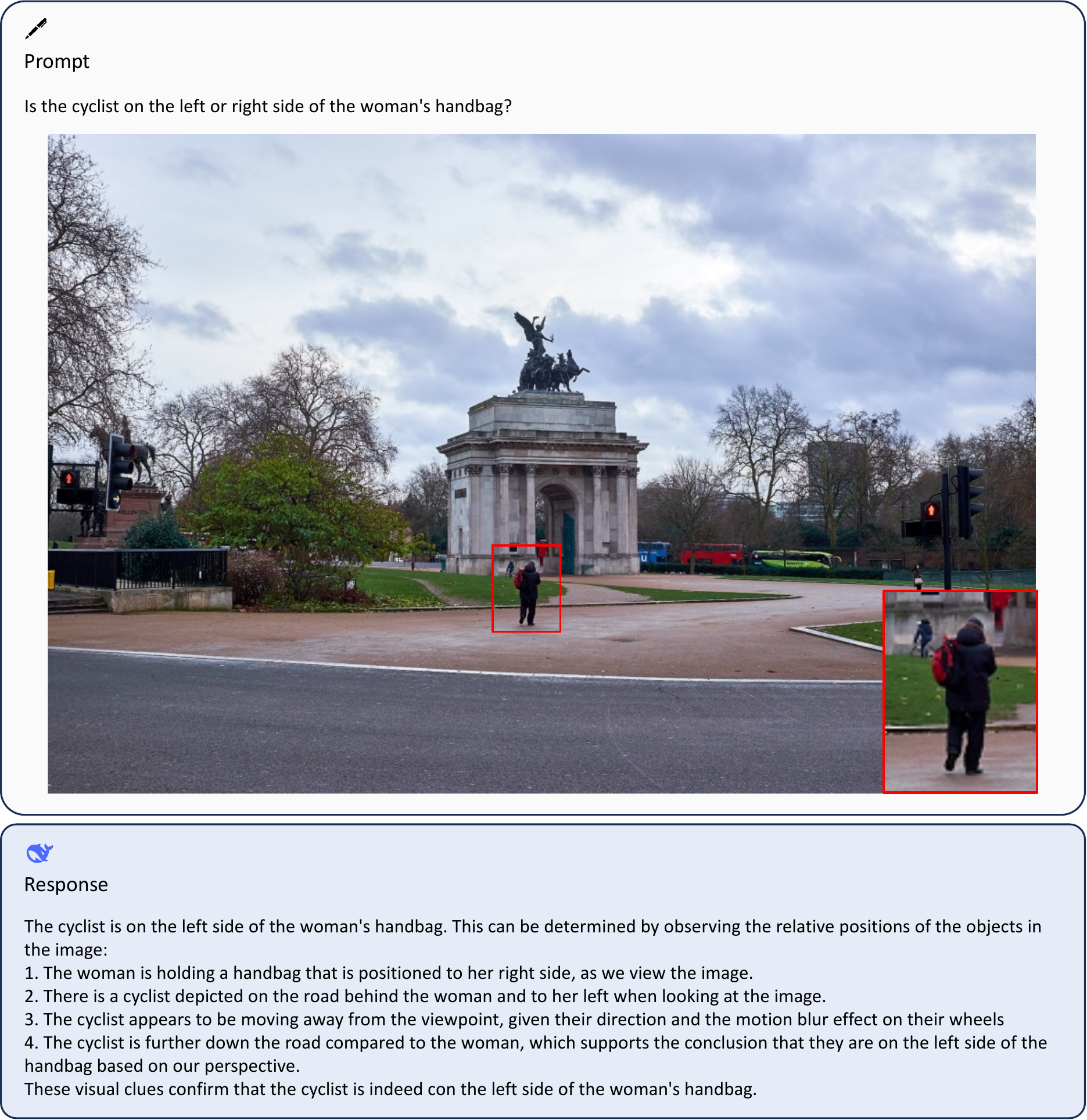}
\caption{Visualization results. 
DeepSeek-VL is capable of capturing tiny object and giving organized explanations.}
\label{fig:visualization_6}
\end{figure}

\textbf{Vision-Language Adaptor.} 
We employ a two-layer hybrid MLP to bridge the vision encoder and the LLM. Initially, distinct single-layer MLPs are used to process high-resolution features and low-resolution features separately. Subsequently, these features are concatenated along their dimensions and then transformed into the LLM's input space through another layer of MLP.

\textbf{Language Model.} Our language model is built upon DeepSeek LLM~\citep{deepseek-llm} whose micro design largely follows the design of LLaMA~\citep{llama, llama2}, adopting a Pre-Norm structure with RMSNorm~\citep{zhang2019root} function and using SwiGLU~\citep{shazeer2020glu} as the activation function for the Feed-Forward Network (FFN), with an intermediate layer dimension of $\frac{8}{3}d_{model}$. It also incorporates Rotary Embedding~\citep{su2024roformer} for positional encoding and uses the same tokenizer with DeepSeek-LLM. We introduce a family of DeepSeek-VL models. 
Given our objective of conducting joint pretraining with multimodal and language, we select an intermediate checkpoint from DeepSeek's pretrained models to continue pretraining.

Specifically, the DeepSeek-VL-1B model is constructed based on the DeekSeek-LLM-1B model, which underwent training with an approximate corpus of 500 billion text tokens. And the DeekSeek-VL-7B model is developed leveraging the DeepSeek-LLM-7B model trained with an estimated 2 trillion text tokens.

        

\begin{figure}[t]
\centering
\includegraphics[width=1.0\textwidth]{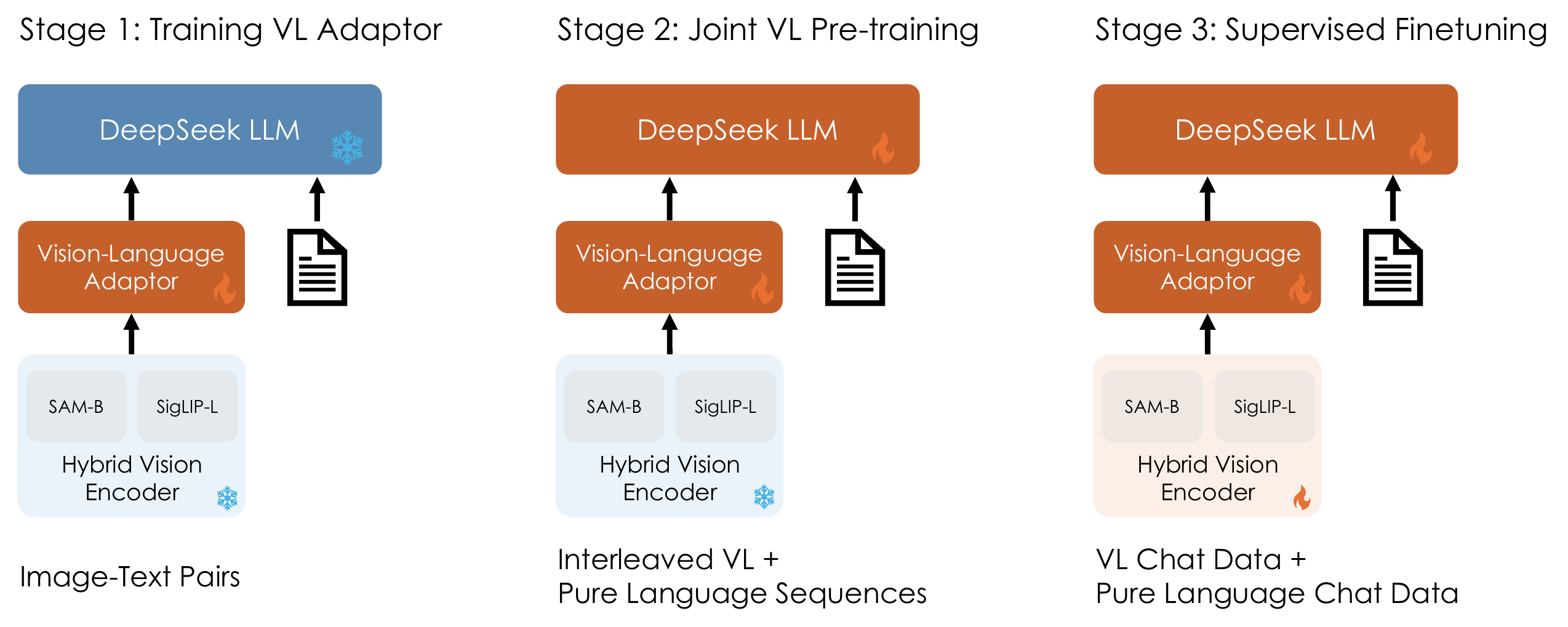}
\caption{Our training pipelines consist of three stages. Stage 1 involves training the Vision-Language (VL) adaptor while keeping the hybrid vision encoder and language model fixed. Stage 2 is the crucial part of the joint vision and language pretraining, where both VL adaptor and language model are trainable. Stage 3 is the supervised fine-tuning phase, during which the low-resolution vision encoder SigLIP-L, VL adaptor, and language model will be trained.}
\label{fig:training_stage}
\end{figure}

\subsection{Training Pipelines}
We train our DeepSeek-VL in three consecutive stages as shown in Figure~\ref{fig:training_stage}: vision-language adaptor warmup, joint vision-language pretraining, and supervised fine-tuning. We currently focus on visual understanding capabilities and only calculate the next token prediction loss on the language part.

\subsubsection{Stage 1: Training Vision-Language Adaptor}

The primary objective of this stage is to establish a conceptual link between visual and linguistic elements within the embedding space, thereby facilitating the comprehensive understanding of depicted entities in the images by the Large Language Model (LLM). Consistent with prior research conducted by LLaVA~\citep{llava-v1} and Instruct-BLIP~\citep{dai2023instructblip}, we adopt a similar approach in which both the vision encoder and the LLM remain frozen during this stage, while solely allowing the trainable parameters within the vision-language adaptor. We utilize a dataset comprising 1.25 million image-text paired captions obtained from ShareGPT4V, along with 2.5 million Document OCR rendering pairs to train the VL adaptor.



Nevertheless, compared to Large Language Models (LLMs), vision-language adaptors (e.g., a 2-layer MLP) have a significantly smaller parameter capacity. This limitation in model capacity restricts the capabilities that can be learned during this stage. A natural question arises: \textbf{Can the law of data scaling be effective at this stage?} To address this question, we conducted a simple experiment in Table~\ref{tab:ablation_stage_1}. The results demonstrate that expanding the data scale at this stage does not provide benefits and may even lead to inferior performance. Consequently, we proceed to unfreeze the Large Language Model (LLM) and investigate efficient vision-language pretraining approaches during stage 2.

\subsubsection{Stage 2: Joint Vision-Language pretraining}

In this stage, we explore effective pretraining strategies which can be considered as an additional stage to enable Large Language Models (LLMs) to comprehend multimodal inputs.
We keep the vision encoder frozen and optimize the language model and VL adaptor.

\begin{figure}[t]
\centering
\includegraphics[width=1.0\textwidth]{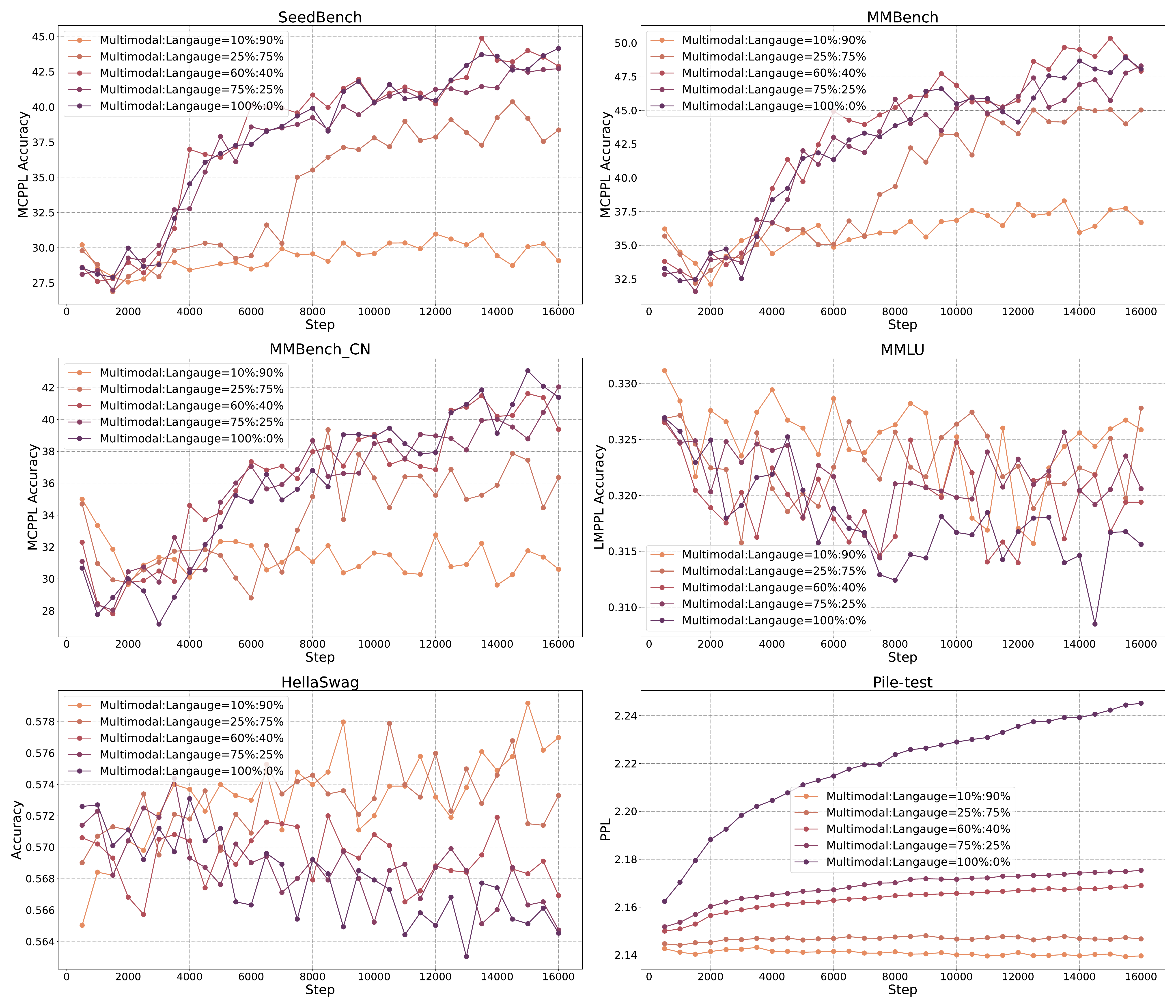}
\caption{Comparative performance results on different modality fusion ratio on training stage 2. An excessively large proportion of multimodal data (multimodal:language=100\%:0\%) leads to significant forgetting of language capabilities in LLMs. A suitable ratio (multimodal:language=70\%:30\%) can effectively mitigate the issue of language forgetting while simultaneously enhancing the model's multimodal abilities. }
\label{fig:ablation_ratio}
\end{figure}

Initially, we attempt to directly train the LLM with multimodal data. However, we find while the metrics for multimodal performance incrementally improved, there is a stark and severe decline in language metrics as illustrated in Figure~\ref{fig:ablation_ratio} (Multimodal:Language=100\%:0\%),. This underscores the inherent challenge in directly conducting multimodal pretraining on the foundation of an LLM, revealing a critical trade-off between enhancing multimodal abilities and preserving linguistic proficiency.

We hypothesize that the observed phenomenon stems from two primary factors: firstly, the majority of multimodal corpora, are overly simplistic and exhibit a significant divergence from the complexity and distribution of linguistic data. Secondly, there appears to be a competitive dynamic between multimodal and linguistic modalities, leading to what can be described as catastrophic forgetting of language capabilities within the LLM.

\noindent\textbf{Joint Language-multimodal Training}
To address this challenge, we devise a straightforward yet effective joint language-multimodal training strategy. During training, we not only engage in multimodal data training but also incorporate a large proportion of language data into the training. This approach aims to balance the training focus, mitigating the adverse effects observed. We conduct experiments on the DeepSeek-VL 1B model in Figure~\ref{fig:ablation_ratio} to explore the impact of varying the modality mixing ratios.

The analysis of the graph yields several key conclusions:
(1). Integrating language data significantly alleviates the decline in language capabilities, demonstrating a substantial improvement in the model's linguistic performance.
(2). The inclusion of language data does not lead to a significant loss in multimodal performance, indicating that the model retains its multimodal processing abilities.
(3). The performance of different modalities is strongly correlated with their respective proportions in the training dataset, substantiating the competitive relationship between the two modalities.
Ultimately, we opt for a training ratio of language to multimodal data of roughly 7:3 for our final model. This ratio enables the model to maintain its language capabilities while simultaneously achieving better pretraining on multimodal data, effectively balancing the development of both language and multimodal proficiencies.


\noindent\textbf{Scaling Vision-Language Pretraining} 
Nevertheless, the pretraining stage of the model incurs a substantial computational cost, and performing iterations on the 7B model requires an excessive amount of computing power and time. One suitable strategy involves conducting experiments on a smaller model, specifically the 1.3B model, and subsequently scaling it up to the 7B model. Fortunately, we have observed that a significant portion of the outcomes obtained from the 1.3B models can be effectively transferred to the 7B model through the utilization of SFT (e.g., the encoder design). However, during the stage 2 training phase, we have encountered considerable fluctuations in the generative metrics of the 1.3B model, rendering it challenging to supervise the training process effectively.
And this has been discussed in ~\cite{schaeffer2024emergent}, "sharp and unpredictable changes might be induced by the researcher’s choice of measurement, even though the model family’s per-token error rate changes smoothly, continuously and predictably with increasing scale." Subsequent experiments have led us to identify the root causes of this issue: the limited capacity of the 1.3B model and the absence of SFT data within the training dataset, both of which hinder the model's ability to accurately follow instructions. Even when the model possesses knowledge of the correct options, it struggles to generate them precisely. 

To mitigate these challenges, we adopte a dual-pronged approach. Firstly, we employ the Multi-choice PPL methodology to monitor the model's progress. This involves inputting not only the prompt and image into the network but also all the answer associated with the question. Subsequently, we calculate the PPL for each answer position (e.g., A, B, C, D) and select the option deemed correct by the model as the final answer. Secondly, we introduce SFT data into the training dataset at a minimal proportion, allowing the model to acquire some proficiency in following instructions. The combination of these two approaches ensures the maintenance of stable training metrics for the 1.3B model and bring better performance after stage3.

\subsubsection{Stage 3: Supervised Fine-tuning}

In this phase, we finetune the pretrained DeepSeek-VL model with instruction-based fine-tuning to bolster its ability to follow instructions and engage in dialogue, culminating in the creation of the interactive DeepSeek-VL-Chat model. 
We optimize the language model, VL adaptor, and hybrid vision encoder with the vision-language SFT data as shown in Table~\ref{tab:sft_datasets}, SAM-B remains frozen due to the limited GPU memory. We only supervise answers and special tokens and mask the system and user prompts.
To guarantee the model's comprehensive proficiency in dialogue, we utilize a blend of multimodal data and pure text dialogue data used in DeepSeek-LLM. This approach ensures the model's versatility across various dialogue scenarios.

\begin{figure}[t!]
\centering
\includegraphics[width=1.0\textwidth]{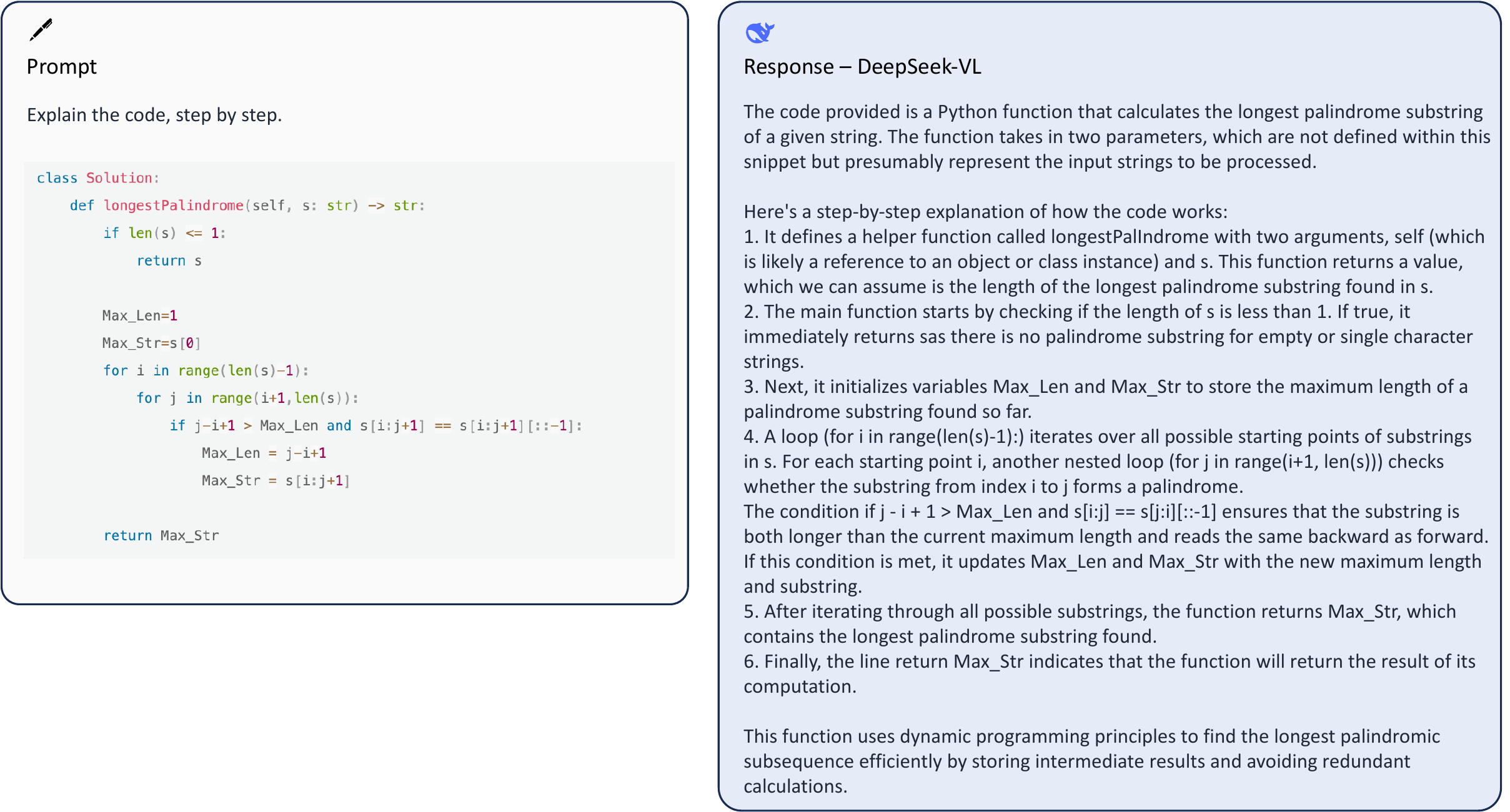}
\caption{Visualization results. 
DeepSeek-VL can understand Python code and provide detailed and organized explanations.}
\label{fig:visualization_7}
\end{figure}

\subsection{Hyperparameters and Infrastructures}
\label{sec:intro-hyper-params}

The detailed hyperparameters of all stages are illustrated in Table~\ref{tab:hyper}. We train and evaluate our DeepSeek-VL with HAI-LLM~\citep{haillm}, a lightweight and efficient distributed training framework. Since we use visual encoders to convert images into embedding vectors and then treat image embeddings and text embeddings uniformly, we can easily adapt pipeline parallelism to VL model training: all we need to do is to view visual encoders and text embedding as a single module and take it as the first layer of the resulting model. This very first layer has a complicated model structure and precludes standard tensor parallelism technique, but luckily it requires relatively small computation compared to upper standard transformer blocks. We therefore simply recompute the visual encoder forward pass in all tensor parallel ranks. The existence of visual encoders also leads to non-uniform execution time across model layers, so we re-divide model layers between pipeline parallelism ranks to achieve better load balance and throughput. The upper layers of DeepSeek-VL are exactly the same as those in DeepSeek-LLM. With such minor modification, we can now perform canonical 3D parallelism techniques as in Megatron~\citep{megatron, megatron2, megatron3} and overlap computation and communication as in DeepSeek-LLM~\citep{deepseek-llm}. 
DeepSeek-VL-7B consumed 5 days on a cluster of 64 nodes, each comprising 8 Nvidia A100 GPUs, while DeepSeek-VL-1B consumed 7 days on a setup involving 16 nodes.

\renewcommand{\arraystretch}{1.1}
\begin{table}[t!]
\centering
\small
\begin{tabular}{l|ccc|ccc}
\toprule
& \multicolumn{3}{c}{DeepSeek-VL 1B} & \multicolumn{3}{c}{DeepSeek-VL-7B} \\
Vision Encoder& \multicolumn{3}{c}{SigLIP} & \multicolumn{3}{c}{SigLIP+SAM} \\
\midrule
\textbf{Hyperparameters} & \textbf{Stage 1} & \textbf{Stage 2}         & \textbf{Stage 3} & \textbf{Stage 1} & \textbf{Stage 2}         & \textbf{Stage 3} \\
\midrule
Learning rate & $1.0\times10^{-3}$   & $3\times10^{-5}$ & $2.0\times10^{-5}$& $1.0\times10^{-3}$   & $4.2\times10^{-5}$ & $2.0\times10^{-5}$  \\
LR scheduler  & Cosine & Step & Cosine & Cosine & Step & Cosine  \\
Weight decay  & 0.0& 0.0& 0.0& 0.0& 0.0& 0.0   \\
Gradient clip & 1.0 & 1.0 & 1.0 & 1.0 & 1.0 & 1.0 \\
Optimizer     & \multicolumn{3}{c}{AdamW($\beta_1=0.9, \beta_2=0.95$)}& \multicolumn{3}{c}{AdamW($\beta_1=0.9, \beta_2=0.95$)} \\
Warm-up steps    & 128      & 2000  & 256 & 128      & 2000  & 256   \\
Training steps   & 15000    & 96000 & 10000& 15000    &  42000 & 10000 \\
Batch size       & 256      & 1024 & 256& 256      & 2304 & 256    \\
Sequence length  & 512      & 4096 & 4096& 512      & 4096 & 4096   \\
Sequence packing & $\times$ & \checkmark & $\times$ & $\times$ & \checkmark & $\times$ \\
Pipeline parallelism   & $\times$ & $\times$ & $\times$ & $\times$ & \checkmark & \checkmark \\
\bottomrule
\end{tabular}
\caption{Detailed hyperparameters of our DeepSeek-VL.}
\label{tab:hyper}
\end{table}

\section{Evaluation}

\subsection{Public Multimodal Benchmarks Evaluation}

We evaluate our models on a series of public benchmarks:

\textbf{Multimodal comprehensive understanding} datasets: MMMU~\citep{yue2023mmmu}, CMMMU~\citep{zhang2024cmmmu}, MMBench~\citep{liu2023mmbench}, MMBench-CN~\citep{liu2023mmbench}, SeedBench~\citep{li2023seed} and MMV~\citep{yu2023mm}. We compare DeepSeek-VL with competitors on MMB/MMC-dev as current official test download link is no longer active.

\textbf{Chart/table understanding} datasets: OCRBench~\citep{liu2023hidden};

\renewcommand{\arraystretch}{1.1}
\begin{table}[t!]
\centering
\small
\scalebox{1}{
\tabcolsep3pt
\begin{tabular}{lcccccccccccc}
\toprule

                   & LLM & MMMU & CMMMU  & MMB & MMC & SEED & OCRB & POPE   & MathV & MMVet \\
\midrule
                   \multicolumn{3}{l}{\textbf{Close-source LMMs}:}
                    \\

                   Gemini Pro & Unk & 48.9& -  & 75.2 & 74.0 & 70.7 & 659 & -& 45.2 & 59.2\\
                    GPT-4V & Unk & 56.8 & 42.5 & 75.0 & 74.7 & 71.6 & 659 & -&  47.8 & 49.9 \\
                    Qwen-VL-Plus & Unk &  45.2& 39.5 & 66.2 & 69.6 & 72.7 &- & - & 43.3 & 55.7 \\
                    Qwen-VL-MAX & Unk &  51.4&- & 78.1 & 76.4 & 72.7 & -& -&  51.0 & 61.8 \\

                   \midrule
                   \multicolumn{3}{l}{\textbf{Open-source 13B LMMs}:}
\\
LLaVA-1.5   & 13B      &    36.4     &    -  &   68.2      &    61.9        &   68.2 & 331 & 85.9 &  26.4 & 38.3   \\
VILA        & 13B      &- &   -          &    70.3     & 64.3           & -&  -& 84.2 & -  & 38.8   \\
LLaVA-Next & 13B & 36.2 & - & 70.0 & 64.4 & 71.9 & - & 86.7 & 35.3 & 48.4\\ \midrule

\multicolumn{3}{l}{\textbf{Open-source 7B LMMs}:} \\

EMU2-Chat      & 7B      &     36.3      &   23.8   &   63.6      &      45.9      &   68.9  &- & - &  30.0 & 31.0\\
Qwen-VL-Chat & 7B & 37.0 & - & 60.6 & 56.7 & 64.8 &  - & - & 33.8 & 47.3\\
CogVLM & 7B &  37.3    &  24.8     &      63.7   &  53.8          &  68.8& -& -  & 34.7 & \bf54.5   \\
LLaVA-Next & 7B & 35.8 & - & 67.4 & 60.0 & 70.2 & - & 86.5 & 34.6 & 43.9\\
Yi-VL & 6B & \bf37.8 & 35.8 & 68.2 & 68.9 & 67.6 & -& -& 28.0 & 31.1 \\
\midrule
DeepSeek-VL (ours) & 7B &   36.6   &  \bf37.9       &    \bf73.2     &    \bf72.8         & \bf70.4 & 456 & \bf88.1 &  \bf36.1 & 41.5 \\   

\bottomrule
\end{tabular}}
\caption{The comparison between different multi-modal models. The top half are proprietary models, while the bottom are open-source models.}
    \label{tab: sota}
\end{table}

\renewcommand{\arraystretch}{1.1}
\begin{table}[t!]
\centering
\small
\scalebox{1}{
\tabcolsep3pt
\begin{tabular}{lcccccccccccc}
\toprule

                   & LLM & MMMU & CMMMU  & MMB & MMC & SEED & OCRB & POPE   & MathV & MMVet \\
\midrule
\textbf{Tiny Model}: \\
MobileVLM & 1.4B &- &- & 53.2 &- &- &- & 84.5 &-&- \\
MobileVLM & 2.7B & -&- & 59.6 &- &- &- & 84.9 &-&-\\
MobileVLM V2 & 1.4B &- &- & 59.6 &- &- &- & 84.3 &-&- \\
MobileVLM V2 & 2.7B & -&- & 63.2 &- &- &- & 84.7 &-&-\\
LLaVA-Phi & 2.7B & -&- & 59.5 & - & -& -& 85.0 & - & 28.9 \\
\midrule
DeepSeek-VL (ours) & 1.3B &   32.2   &  27.4       &    \bf64.6    &    61.3        & 66.7 & 409 & \bf87.6 &  31.1 & \bf34.8 \\   
 
\bottomrule
\end{tabular}}
\caption{The comparison between tiny multi-modal models.}
    \label{tab: sota_tiny}
\end{table}

\textbf{Hallucination} datasets: POPE~\citep{li2023evaluating};

\textbf{Scientific problem} datasets: ScienceQA~\citep{lu2022learn} and MathVista~\citep{lu2023mathvista}.

We apply generation-based evaluation with greedy decoding. The generation-based evaluation here refers to letting the model generate free texts and parsing results from generated texts.
The comparative results, as illustrated in Table~\ref{tab: sota}, show that DeepSeek-VL-7B surpasses most open-source models of similar size across a wide range of benchmarks.

DeepSeek-VL outperforms open-source models of similar size in benchmarks such as MMB, MMC, and SEEDbench, even approaching proprietary models (DeepSeek-VL vs. GPT-4V = 70.4 vs. 71.6 on seedbench), demonstrating its powerful natural image comprehension capability. The model also surpasses all open-source models in mathematical logic, but still lags significantly behind proprietary models like GPT-4V (36.1 vs. 47.8 on MathVista). This difference could be attributed to the variance in base model sizes.

Furthermore, as shown in Table~\ref{tab: sota_tiny}, DeepSeek-VL-1.3B significantly outperforms models of comparable size. It demonstrates superior performance compared to leading open-source models in the MMB benchmark test, while utilizing only close to half the parameters (1.3B vs. 2.7B), indicating its robust natural image comprehension capability. DeepSeek-VL-1.3B even achieves comparable results to 7B open-source models on MathVista, further validating the powerful logical understanding capabilities of the DeepSeek-VL family.

\subsection{Public Language Benchmarks Evaluation}

We evaluate our models on the following public language benchmarks:

\textbf{Multi-subject multiple-choice} datasets including MMLU~\citep{mmlu}.

\textbf{Language understanding and reasoning} datasets including HellaSwag~\citep{hellaswag}.

\textbf{Language modeling} datasets including Pile~\citep{pile}.

\renewcommand{\arraystretch}{1.1}
\begin{table}[t!]
\centering
\small
  \scalebox{1}{
\tabcolsep13pt
\begin{tabular}{c|c|c|c|c}
\toprule
\multirow{2}{*}{} & \multirow{2}{*}{Version} & DeepSeek-VL  & DeepSeek-VL & DeepSeek-LLM \\
 &  & 1B Chat  & 7B Chat  & 7B Chat\\
 & Encoder & SigLIP  & SigLIP+SAM  & None \\
\midrule
\multirow{6}{*}{Benchmark} & HellaSwag & 56.0  & 68.4 & \bf68.5 \\

&MMLU  & 32.5 &  \bf52.4 &  49.4 \\

&GSM8K  & 18.0 &  55.0 &  \bf63.0 \\

&MBPP  & 10.0  & \bf35.2 &  \bf35.2 \\

&AGIEval  & 14.0  & \bf27.8 & 19.3  \\

\bottomrule
\end{tabular}}
    \caption{The performance on language benchmarks.  }
    \label{tab: chat}
\end{table}

\textbf{Math} datasets including GSM8K~\citep{gsm8k}.

\textbf{Code} datasets including  MBPP~\citep{mbpp}.

\textbf{Standardized exams} including AGIEval~\citep{agieval}.

We apply perplexity-based evaluation to datasets that require answers to be chosen from several options. These datasets include HellaSwag and MMLU. The perplexity-based evaluation here refers to calculating the perplexity of each option and selecting the lowest one as the model prediction. Perplexity-based evaluation helps to distinguish subtle probability difference between model predictions and avoids discontinuity of exact match style evaluation. We apply generation-based evaluation with greedy decoding for GSM8K and AGIEval. The generation-based evaluation here refers to letting the model generate free texts and parsing results from generated texts. We apply language-modeling-based evaluation for Pile-test, which means calculating the bits-per-byte on the test corpus. And the results are illustrated in Table~\ref{tab: chat}

It can be observed that across the majority of language benchmarks, DeepSeek-VL performs comparably to, or even surpasses, DeepSeek-7B. For instance, it achieves scores of 68.4 vs. 68.5 on HellaSwag, which serves as a general benchmark for evaluating general language ability. DeepSeek-VL outperforms DeepSeek-7B on metrics such as MMLU and AGIEval, indicating that multimodal training methods may even aid in language tasks. Nevertheless, DeepSeek-VL-7B shows a certain degree of decline in mathematics (GSM8K), which suggests that despite efforts to promote harmony between vision and language modalities, there still exists a competitive relationship between them. This could be attributed to the limited model capacity (7B), and larger models might alleviate this issue significantly. Overall, DeepSeek-VL strives to achieve the goal of minimizing declines in language capability while addressing these challenges.

\subsection{Human Evaluation}

To further explore the capabilities of our DeepSeek-VL, we independently construct a dataset for manual evaluation. This dataset comprises 100 questions, divided into seven categories, each encompassing specific tasks. These categories and tasks are same as our taxonomy for the in-house SFT data, as shown in Table~\ref{tab:image_understanding_taxonomy}. This approach ensures that the tasks we test are universal and encompass the majority of use cases for multimodal models.

Moreover, based on the categories and tasks described in existing reports, we collect similar image materials and developed prompts. The sources for these image materials include royalty-free image communities and photographs taken by the researchers. This methodical collection and prompt formulation process ensures our dataset is both comprehensive and representative of real-world multimodal model applications.

\begin{figure}[t!]
\centering
\includegraphics[width=1.0\textwidth]{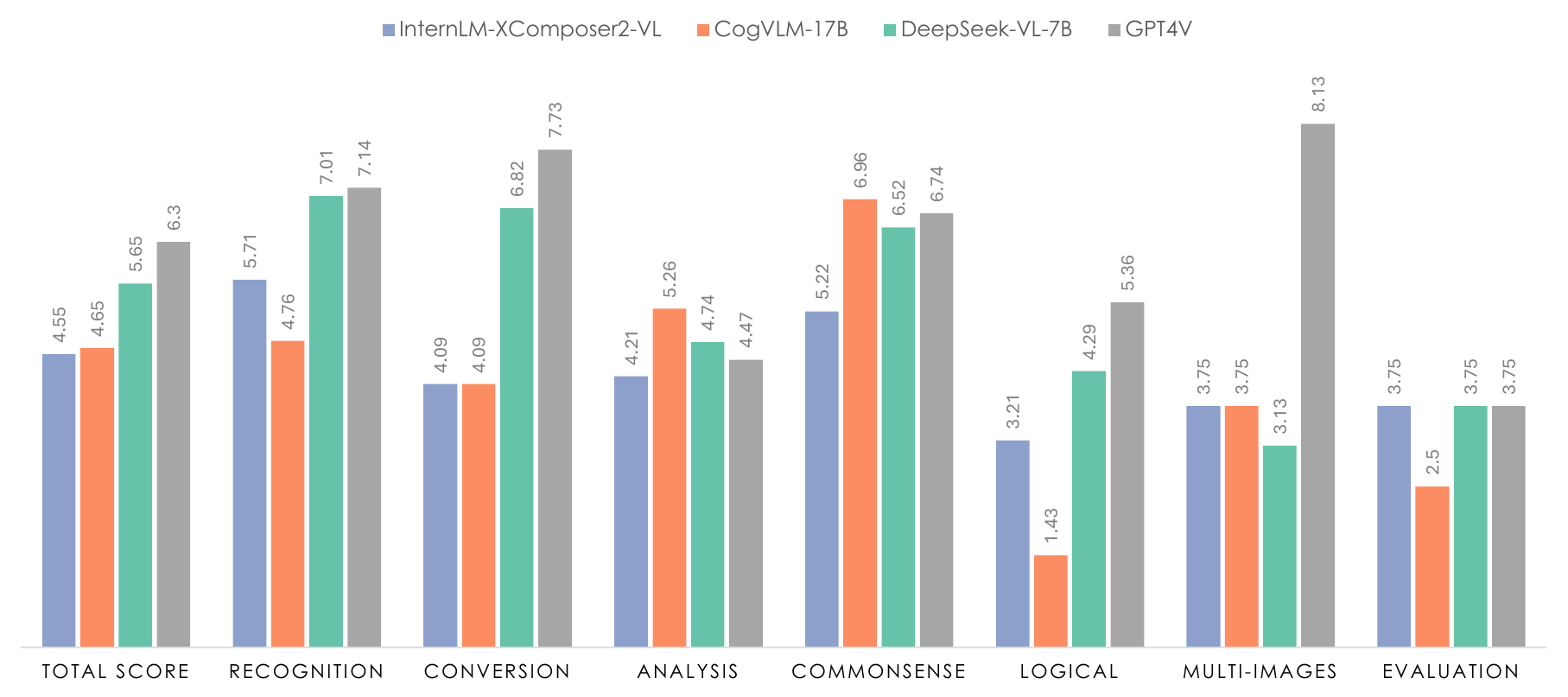}
\caption{Human evaluation results on InternLM-XComposer2-VL~\citep{dong2024internlm}, CogVLM~\citep{cogvlm}, DeepSeek-VL and GPT-4V~\citep{gpt4v}.}
\label{fig:human_eval}
\end{figure}

We compare our DeepSeek-VL-7B with InternLM-XComposer2-VL, CogVLM and GPT-4V as shown in Figure~\ref{fig:human_eval} (and we also provide visualization results in Appendix~\ref{sec:appendix}). 
GPT-4V demonstrates exceptional performance across most dimensions. All open-source models are still far behind GPT-4V in logical reasoning, highlighting the necessity of scaling up the size of Large Language Models (LLMs). DeepSeek-VL-7B achieves better results in overall performance, reaching outcomes close to GPT-4V in Recognition, Conversion, and Commonsense Reasoning.

\begin{figure}[t!]
\centering
\includegraphics[width=0.9\textwidth]{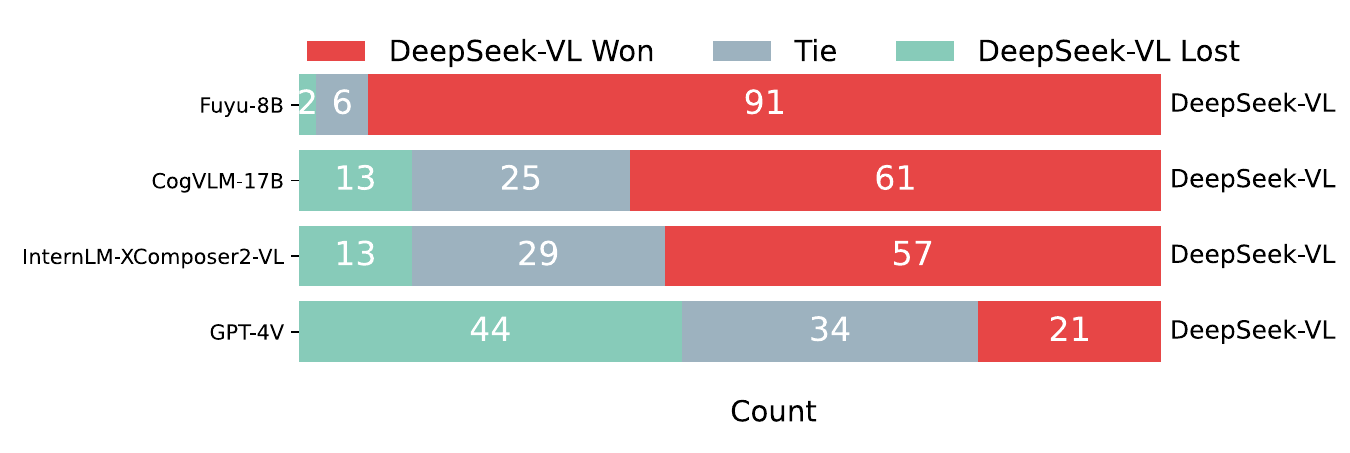}
\caption{GPT-4V-based Evaluation Results of DeepSeek-VL vs. Other Models: The chart depicts results from a GPT-4V-based assessment across 99 test samples, demonstrating DeepSeek-VL's favorable outcomes against both open-source and proprietary models.}
\label{fig:win-rate}
\end{figure}

In addition, we conduct a comparative assessment using GPT-4V to evaluate the performance of DeepSeek-VL against other models across a set of 99 test samples designed for human evaluation.
Following \citep{zheng2024judging}, we show GPT-4V the question and the answers from two different models and ask GPT-4V to determine which one is better or declare a tie.
The results indicate a preference for DeepSeek-VL's responses in the majority of cases, as GPT-4V tends to rate the quality of DeepSeek-VL's answers more favorably. 
As illustrated in Figure~\ref{fig:win-rate}, DeepSeek-VL is judged to be superior in over 60\% of instances when compared to open-source multimodal models, including Fuyu-8B, CogVLM-17B, and InternLM-XComposer2-VL. 
Moreover, in comparison with other proprietary models, such as GPT-4V itself, DeepSeek-VL demonstrates comparably exceptional performance.

\subsection{Ablation Study}
\label{sec:ablation}

\noindent\textbf{Scale Up Projector Training}
We expand the dataset for stage 1 (projector warmup) and subsequently apply supervised fine-tuning. The results, depicted in Figure~\ref{tab:ablation_stage_1}, demonstrate that augmenting the training data volume does not enhance performance at this stage. This implies that the projector's capacity is inherently constrained, rendering it incapable of capturing the extensive knowledge necessary for multimodal tasks.

\renewcommand{\arraystretch}{1.1}
\begin{table}[t!]
\centering
\small
\scalebox{1.0}{
\tabcolsep13pt
\begin{tabular}{ccccccc}
\toprule

     Stage 1, Training Step              & MMB & MMC & SEED  & POPE    & MMMU & Average \\

                   \midrule
                   2K & \bf59.0 & 54.0 & \bf61.8 & 82.3  & \bf30.3 & \bf57.5 \\
                   8K & 58.0 & 45.0 & 58.5 & \bf84.9  & 29.2 & 55.1\\
                   20K & 56.0 & 52.3 & 59.0 & 81.7  & 28.6 & 55.5\\
                   80K & 58.1 & \bf55.0 & 58.6 & 78.6  & 27.9 & 55.6\\
\bottomrule
\end{tabular}
}
\caption{Comparative directly SFT performance results on scaling up stage 1 data. The results demonstrate that expanding the data scale at this stage does not yield benefits, or even results in worse performance.}
    \label{tab:ablation_stage_1}
\end{table}

\renewcommand{\arraystretch}{1.1}
\begin{table}[t!]
\centering
\small
\scalebox{1}{
\tabcolsep10pt
\begin{tabular}{ccccccccc}
\toprule

     Stage 1 & Stage 2 & Stage 3              & MMB & MMC & SEED  & POPE    & MMMU & Average\\

                   \midrule
                    \checkmark&  & \checkmark & 59.4 & 54.2 & 61.4 & 82.5  & 29.2 & 57.4 \\
                   &\checkmark &\checkmark & 63.4 & 60.5 & 65.9 & 87.1  & 31.8 & 61.7 \\
                   \checkmark&\checkmark &\checkmark & 64.3 & 61.3 & 66.7 & 87.6  & 32.2 & \bf62.4 \\
\bottomrule
\end{tabular}
}
\caption{Analysis of model performance across training stages.}
    \label{tab: ablation_stage}
\end{table}

\noindent\textbf{Training Stage} 
In Table~\ref{tab: ablation_stage}, we examine the contributions of each stage to the model's performance. It's evident that combining stage 1, stage 2, and stage 3 yields significantly better results across all metrics compared to combining stage 1 and stage 3 alone, demonstrating the effectiveness of multimodal pretraining. Additionally, the combination of stage 2 and stage 3 still slightly lags behind the combined performance of stage 1, stage 2, and stage 3, indicating that vision-language adaptor warmup stage remains meaningful.

\noindent\textbf{Modality Group Training} When mixing language and multimodal data, we observe that directly blending them at the batch level significantly reduces training efficiency. This inefficiency arises because each batch gradient backpropagation process waits for the slowest sample to complete. As a result, the predominantly faster-to-process pure language data ends up waiting for the multimodal samples to finish, leading to a decrease in overall training efficiency.

\begin{figure}[t!]
\centering
\includegraphics[width=1.0\textwidth]{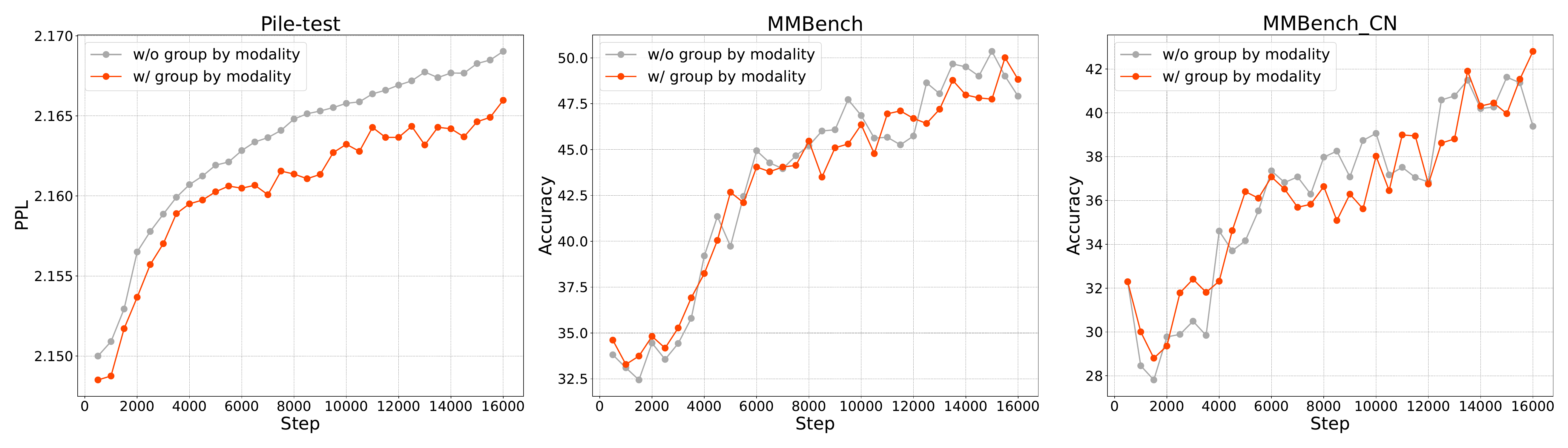}
\caption{Comparative analysis of modality warmup on language (Pile-test) and multimodal (MMBench and MMBench$\_$CN) benchmarks demonstrates that modality grouping consistently surpasses the non-grouped modality approach in language tasks, while simultaneously preserving performance on multimodal tasks on training stage 2 (Multimodal:Language=60\%:40\%). }
\label{fig:ablation_split}
\end{figure}

To address this issue, we experiment with grouping different modalities of data at each global step, sampling distinct modalities separately. This approach involves organizing the training data so that batches are composed either entirely of language data or entirely of multimodal data at different training steps, rather than mixing them within the same batch.

The results are shown in Figure~\ref{fig:ablation_split},
we observe that this method does not compromise the model's performance while enhancing the model's training efficiency by 20\%. This strategy effectively circumvents the bottleneck caused by the disparate processing times between modalities, optimizing the training workflow.

\noindent\textbf{Modality Warmup} Considering that our approach involves multimodal training on the foundation of a language model, directly mixing multimodal data in a fixed proportion from the outset can destabilize the model. To counteract this issue, we propose a simple yet effective modality warm-up strategy. Initially, we set the language data ratio to 1, and then gradually decrease it to the target ratio for the final model training (e.g., 0.7). 

\begin{figure}[t!]
\centering
\includegraphics[width=1.0\textwidth]{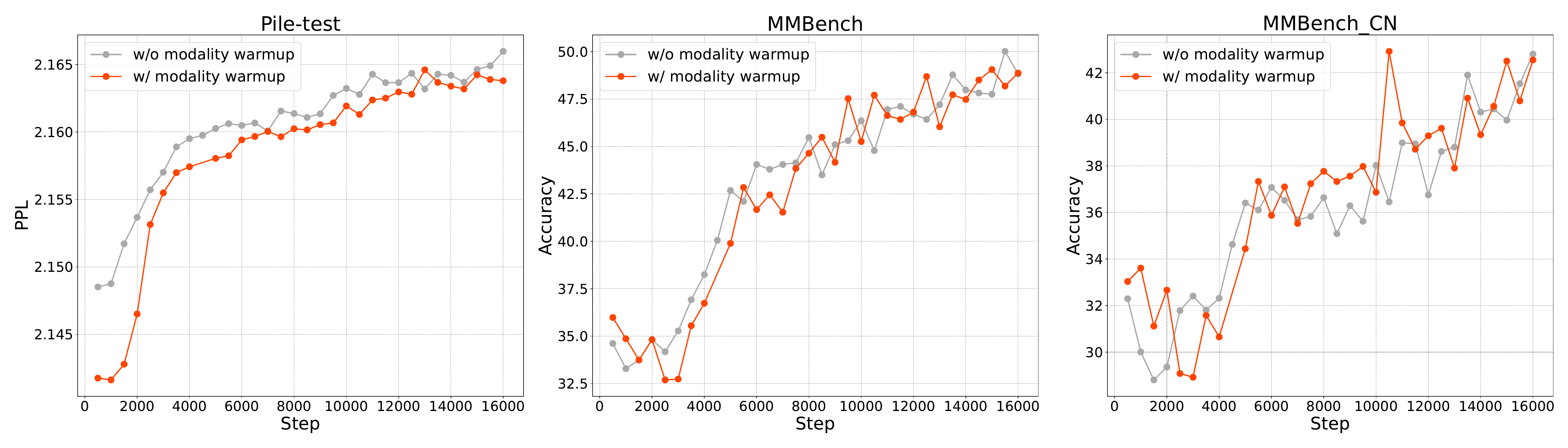}
\caption{Comparative performance results on language (Pile-test) and multimodal (MMBench and MMBench\_CN) benchmarks for modality warmup. Modality warmup consistently matches or surpasses the performance of approaches without modality warmup across all evaluated tasks on training stage 2 (Multimodal:Language=60\%:40\%).}
\label{fig:ablation_warmup}
\end{figure}

Our experiments, as illustrated in Figure~\ref{fig:ablation_warmup}, demonstrate that this strategy effectively prevents a significant decline in language capabilities at the beginning of training, while also yielding comparatively superior outcomes in the final phases for both the language and multimodal domains. This gradual adaptation enables the model to more seamlessly adjust to the incorporation of multimodal data, thereby improving overall training stability and performance.

\begin{figure}[t!]
\centering
\includegraphics[width=0.7\textwidth]{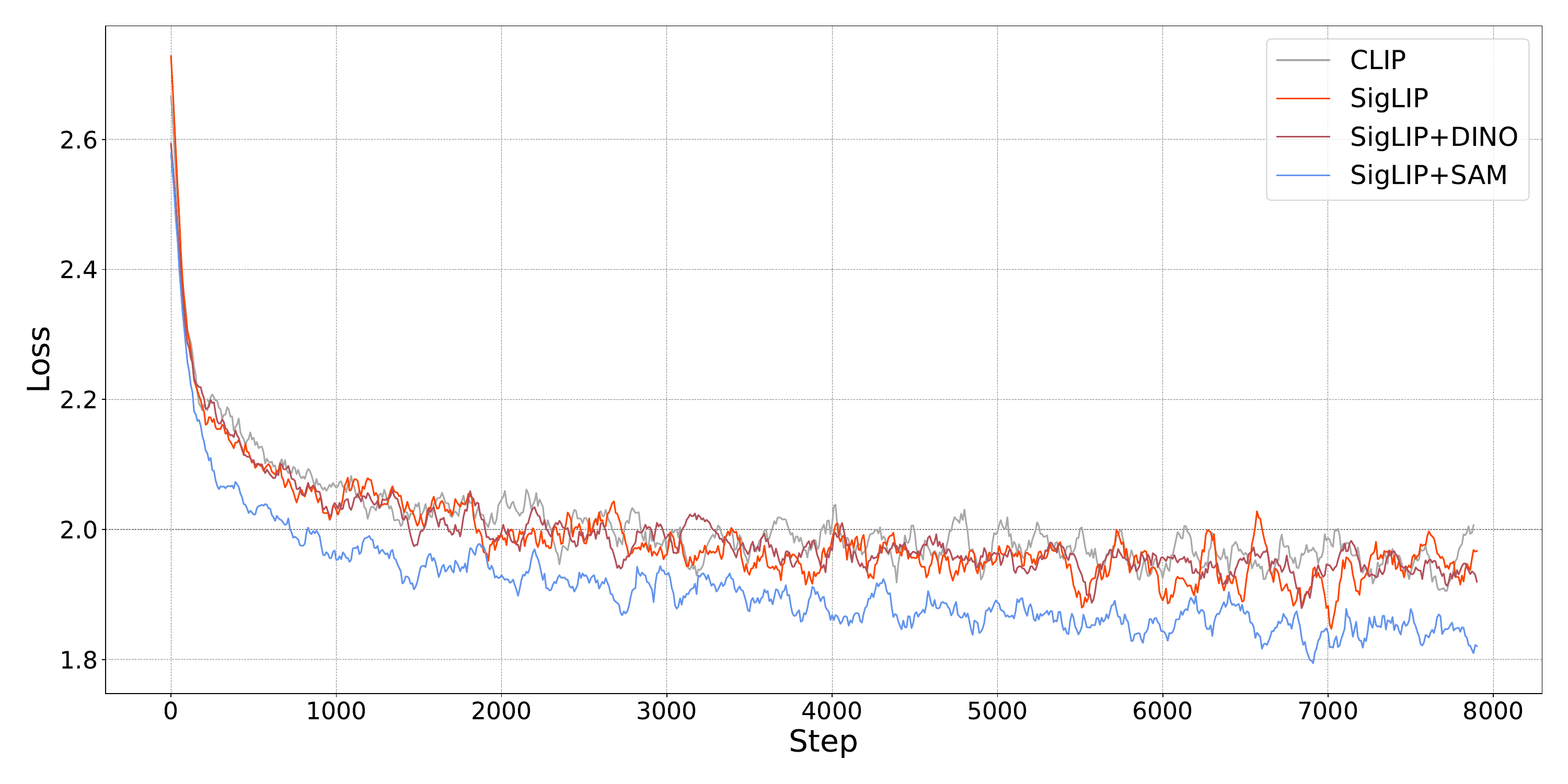}
\caption{Comparative analysis of different vision encoders on training losses in stage 2.}
\label{fig:ablation_vision}
\end{figure}

\noindent\textbf{Vision Encoder Selection} 
In order to better acquire and utilize image information, we compare the training loss of different vision encoders under our training settings except for reducing training steps of stage 2 to 8000 for efficiency. As illustrated in 
Figure~\ref{fig:ablation_vision}, the incorporation of vision-only self-supervised encoders has been found to significantly enhance performance on training loss. To more effectively process high-resolution images, our research ultimately adopts a hybrid vision encoder strategy, combining SigLIP with SAM for our model's implementation.

\noindent\textbf{Vision-Language Adaptor Design}
To improve the efficiency of extracting information from the visual encoder while adhering to current token length constraints, adjustments can be made to the Vision-Language adaptor in two main ways: the method used to combine visual features and the design of the MLP adaptor.

Previous studies \citep{concat-dino} have indicated that combining visual features along the sequence dimension can lead to better model performance, although this comes with the trade-off of increased computational requirements due to a longer sequence of visual feature tokens. As demonstrated in the top section of Table~\ref{tab:ablation_projecter}, reducing the sequence length by stacking visual features along the image's width or height dimensions before sequence concatenation, in order to keep the sequence length constant, does not achieve better results compared to simply merging them along the embedding dimension in most metrics.
In terms of the adaptor architecture, employing separate MLP adaptors for each vision feature encoder allows for more precise adjustments to the specific values and distribution patterns of visual features, facilitating smoother model training. Conversely, using a shared MLP adaptor for different vision encoders contributes to adequate feature fusion. We adopt a mixed strategy and report stable and improved performance, as outlined in the lower section of Table~\ref{tab:ablation_projecter}.

\renewcommand{\arraystretch}{1.1}
\begin{table}[t!]
\centering
\small
\scalebox{1}{
\tabcolsep6pt
\begin{tabular}{lcccccccc}
\toprule

Architecture & MMB & MMC & SEED  & POPE  & ScienceQA  & MMMU & OCRB & Average \\
                   \midrule
\multicolumn{8}{l}{\textbf{Sequence Concatenation:}}\\
\midrule
Token Pooling - W & 61.2 & \underline{59.6} & 61.6 & 86.5 & \bf57.7 &  \underline{31.6} & 304 & \underline{55.5} \\
Token Pooling - H & 59.9 & 58.3 & 61.6 & 83.8 & 55.0 & \bf32.0 & 291&54.2 \\
\midrule
\multicolumn{8}{l}{\textbf{Embedding Concatenation:}}\\
\midrule
Hybrid MLP & \underline{61.7} & \bf60.1 & \underline{62.9} & \bf87.8 & \underline{56.6} & 31.3 & \underline{309} & \bf55.9\\
Shared MLP & \bf62.0 & 58.9 & 62.5 & \underline{86.6} & 54.7 & 30.2 & \bf318 &55.2\\
Separate MLP & 57.5 & 58.7 & \bf63.1 & 86.5 &  \underline{56.6} & 29.0 & 299 &54.5\\

\bottomrule

\end{tabular}
}
\caption{Comparison of different adaptor architectures using SigLIP and SAM as hybrid vision encoder, Hybrid MLP are used for sequence concatenation experiments. \textbf{Bolded} entries represent the best results, while \underline{underlined} entries denote the second-best results. For calculating the average score, we divide the OCRBench by the total number of questions.}
    \label{tab:ablation_projecter}
\end{table}

\section{Conclusion, Limitation, and Future Work}

In this technical report, we have introduced DeepSeek-VL, a series of Multimodal Large Language Models, available in scales of 1.3B and 6.7B parameters. This report has unveiled the limitations inherent in the predominant projector-based pretraining methodologies, setting the stage for the innovative approach adopted by DeepSeek-VL. By prioritizing a joint vision and language (VL) pretraining phase, DeepSeek-VL transcends traditional models by ensuring that the integration of multimodal data does not compromise the linguistic capabilities of the Large Language Models (LLMs). This is achieved through a strategic warm-up data ratio and the introduction of a hybrid vision encoder, which together enable the efficient processing of high-resolution images without losing sight of semantic richness.

The incorporation of a hybrid vision encoder, capable of handling 1024 x 1024 images within a constrained token budget, underscores our commitment to preserving the nuanced details and semantic integrity across diverse tasks. As a result, DeepSeek-VL emerges as a pioneering model that not only meets but exceeds the standards set by generalist models in its class. It showcases exceptional performance across a wide range of visually-centric benchmarks while sustaining formidable proficiency in language-centric evaluations.

In making DeepSeek-VL publicly available, we aim to catalyze further innovation and exploration within the research community, providing a robust foundation upon which future studies can build. This gesture of openness is intended to facilitate the collective advancement of our understanding and capabilities in handling multimodal data.

Looking ahead, we are excited to announce plans to scale up DeepSeek-VL to larger sizes, incorporating Mixture of Experts (MoE) technology. This forthcoming expansion promises to further enhance the model's efficiency and effectiveness, opening up new horizons for research and application in the field of AI. 

\bibliography{main}

\begin{thebibliography}{90}
\providecommand{\natexlab}[1]{#1}
\providecommand{\url}[1]{\texttt{#1}}
\expandafter\ifx\csname urlstyle\endcsname\relax
  \providecommand{\doi}[1]{doi: #1}\else
  \providecommand{\doi}{doi: \begingroup \urlstyle{rm}\Url}\fi

\bibitem[{01-ai}(2024)]{yi-vl-34b}
{01-ai}.
\newblock Yi-34{B} vision language model.
\newblock \url{https://huggingface.co/01-ai/Yi-VL-34B}, 2024.

\bibitem[Abi(2024)]{screen-to-code}
Abi.
\newblock Screenshot to code.
\newblock \url{https://github.com/abi/screenshot-to-code}, 2024.

\bibitem[{Anna's Archive}(2024)]{annas-archive}
{Anna's Archive}.
\newblock Anna's archive.
\newblock \url{https://annas-archive.org/}, 2024.

\bibitem[Anthropic(2023)]{claude}
Anthropic.
\newblock Introducing {Claude}, 2023.
\newblock URL \url{https://www.anthropic.com/index/introducing-claude}.

\bibitem[Austin et~al.(2021)Austin, Odena, Nye, Bosma, Michalewski, Dohan, Jiang, Cai, Terry, Le, et~al.]{mbpp}
J.~Austin, A.~Odena, M.~Nye, M.~Bosma, H.~Michalewski, D.~Dohan, E.~Jiang, C.~Cai, M.~Terry, Q.~Le, et~al.
\newblock Program synthesis with large language models.
\newblock \emph{arXiv preprint arXiv:2108.07732}, 2021.

\bibitem[Bai et~al.(2023)Bai, Bai, Yang, Wang, Tan, Wang, Lin, Zhou, and Zhou]{qwen-vl}
J.~Bai, S.~Bai, S.~Yang, S.~Wang, S.~Tan, P.~Wang, J.~Lin, C.~Zhou, and J.~Zhou.
\newblock Qwen-vl: A versatile vision-language model for understanding, localization, text reading, and beyond.
\newblock \emph{arXiv preprint arXiv:2308.12966}, 2023.

\bibitem[Bavishi et~al.(2023)Bavishi, Elsen, Hawthorne, Nye, Odena, Somani, and Ta\c{s}\i{}rlar]{fuyu-8b}
R.~Bavishi, E.~Elsen, C.~Hawthorne, M.~Nye, A.~Odena, A.~Somani, and S.~Ta\c{s}\i{}rlar.
\newblock Introducing our multimodal models, 2023.
\newblock URL \url{https://www.adept.ai/blog/fuyu-8b}.

\bibitem[Blecher(2024)]{latex-ocr}
L.~Blecher.
\newblock Latex-ocr.
\newblock GitHub repository, 2024.
\newblock URL \url{https://github.com/lukas-blecher/LaTeX-OCR}.

\bibitem[Blecher et~al.(2023)Blecher, Cucurull, Scialom, and Stojnic]{nougat}
L.~Blecher, G.~Cucurull, T.~Scialom, and R.~Stojnic.
\newblock Nougat: Neural optical understanding for academic documents.
\newblock \emph{arXiv preprint arXiv:2308.13418}, 2023.

\bibitem[Burns et~al.(2023)Burns, Srinivasan, Ainslie, Brown, Plummer, Saenko, Ni, and Guo]{burns2023wiki}
A.~Burns, K.~Srinivasan, J.~Ainslie, G.~Brown, B.~A. Plummer, K.~Saenko, J.~Ni, and M.~Guo.
\newblock A suite of generative tasks for multi-level multimodal webpage understanding.
\newblock In \emph{The 2023 Conference on Empirical Methods in Natural Language Processing (EMNLP)}, 2023.
\newblock URL \url{https://openreview.net/forum?id=rwcLHjtUmn}.

\bibitem[Carter(2024)]{textocr-gpt4v}
J.~Carter.
\newblock Textocr-gpt4v.
\newblock \url{https://huggingface.co/datasets/jimmycarter/textocr-gpt4v}, 2024.

\bibitem[Chen et~al.(2023)Chen, Li, Dong, Zhang, He, Wang, Zhao, and Lin]{sharegpt4v}
L.~Chen, J.~Li, X.~Dong, P.~Zhang, C.~He, J.~Wang, F.~Zhao, and D.~Lin.
\newblock Sharegpt4v: Improving large multi-modal models with better captions.
\newblock \emph{arXiv preprint arXiv:2311.12793}, 2023.

\bibitem[Chng et~al.(2019)Chng, Liu, Sun, Ng, Luo, Ni, Fang, Zhang, Han, Ding, et~al.]{chng2019icdar2019}
C.~K. Chng, Y.~Liu, Y.~Sun, C.~C. Ng, C.~Luo, Z.~Ni, C.~Fang, S.~Zhang, J.~Han, E.~Ding, et~al.
\newblock Icdar2019 robust reading challenge on arbitrary-shaped text-rrc-art.
\newblock In \emph{2019 International Conference on Document Analysis and Recognition (ICDAR)}, pages 1571--1576. IEEE, 2019.

\bibitem[Cobbe et~al.(2021)Cobbe, Kosaraju, Bavarian, Chen, Jun, Kaiser, Plappert, Tworek, Hilton, Nakano, et~al.]{gsm8k}
K.~Cobbe, V.~Kosaraju, M.~Bavarian, M.~Chen, H.~Jun, L.~Kaiser, M.~Plappert, J.~Tworek, J.~Hilton, R.~Nakano, et~al.
\newblock Training verifiers to solve math word problems.
\newblock \emph{arXiv preprint arXiv:2110.14168}, 2021.

\bibitem[Dai et~al.(2023)Dai, Li, Li, Tiong, Zhao, Wang, Li, Fung, and Hoi]{dai2023instructblip}
W.~Dai, J.~Li, D.~Li, A.~M.~H. Tiong, J.~Zhao, W.~Wang, B.~Li, P.~Fung, and S.~Hoi.
\newblock Instructblip: Towards general-purpose vision-language models with instruction tuning, 2023.

\bibitem[DeepSeek-AI(2024)]{deepseek-llm}
DeepSeek-AI.
\newblock Deepseek llm: Scaling open-source language models with longtermism.
\newblock \emph{arXiv preprint arXiv:2401.02954}, 2024.
\newblock URL \url{https://github.com/deepseek-ai/DeepSeek-LLM}.

\bibitem[Dong et~al.(2024)Dong, Zhang, Zang, Cao, Wang, Ouyang, Wei, Zhang, Duan, Cao, et~al.]{dong2024internlm}
X.~Dong, P.~Zhang, Y.~Zang, Y.~Cao, B.~Wang, L.~Ouyang, X.~Wei, S.~Zhang, H.~Duan, M.~Cao, et~al.
\newblock Internlm-xcomposer2: Mastering free-form text-image composition and comprehension in vision-language large model.
\newblock \emph{arXiv preprint arXiv:2401.16420}, 2024.

\bibitem[{echo840}(2024)]{detailed_caption}
{echo840}.
\newblock Detailed caption dataset.
\newblock \url{https://huggingface.co/datasets/echo840/Detailed_Caption}, 2024.

\bibitem[Foundation()]{wikidump}
W.~Foundation.
\newblock Wikimedia downloads.
\newblock URL \url{https://dumps.wikimedia.org}.

\bibitem[Gao et~al.(2023)Gao, Pi, Zhang, Ye, Zhong, Wang, Hong, Han, Xu, Li, et~al.]{geo170k}
J.~Gao, R.~Pi, J.~Zhang, J.~Ye, W.~Zhong, Y.~Wang, L.~Hong, J.~Han, H.~Xu, Z.~Li, et~al.
\newblock G-llava: Solving geometric problem with multi-modal large language model.
\newblock \emph{arXiv preprint arXiv:2312.11370}, 2023.

\bibitem[Gao et~al.(2020)Gao, Biderman, Black, Golding, Hoppe, Foster, Phang, He, Thite, Nabeshima, et~al.]{pile}
L.~Gao, S.~Biderman, S.~Black, L.~Golding, T.~Hoppe, C.~Foster, J.~Phang, H.~He, A.~Thite, N.~Nabeshima, et~al.
\newblock The {Pile}: An {800GB} dataset of diverse text for language modeling.
\newblock \emph{arXiv preprint arXiv:2101.00027}, 2020.

\bibitem[Google(2023)]{bard}
Google.
\newblock An important next step on our {AI} journey, 2023.
\newblock URL \url{https://blog.google/technology/ai/bard-google-ai-search-updates/}.

\bibitem[Hendrycks et~al.(2020)Hendrycks, Burns, Basart, Zou, Mazeika, Song, and Steinhardt]{mmlu}
D.~Hendrycks, C.~Burns, S.~Basart, A.~Zou, M.~Mazeika, D.~Song, and J.~Steinhardt.
\newblock Measuring massive multitask language understanding.
\newblock \emph{arXiv preprint arXiv:2009.03300}, 2020.

\bibitem[High-flyer(2023)]{haillm}
High-flyer.
\newblock Hai-llm: 高效且轻量的大模型训练工具, 2023.
\newblock URL \url{https://www.high-flyer.cn/en/blog/hai-llm}.

\bibitem[Hsiao et~al.(2022)Hsiao, Zubach, Wang, et~al.]{screenqa}
Y.-C. Hsiao, F.~Zubach, M.~Wang, et~al.
\newblock Screenqa: Large-scale question-answer pairs over mobile app screenshots.
\newblock \emph{arXiv preprint arXiv:2209.08199}, 2022.

\bibitem[Hu et~al.(2023)Hu, Shi, Xu, Ye, Ye, Yan, Li, Qian, Zhang, and Huang]{m-paper}
A.~Hu, Y.~Shi, H.~Xu, J.~Ye, Q.~Ye, M.~Yan, C.~Li, Q.~Qian, J.~Zhang, and F.~Huang.
\newblock mplug-paperowl: Scientific diagram analysis with the multimodal large language model.
\newblock \emph{arXiv preprint arXiv:2311.18248}, 2023.

\bibitem[HuggingFaceM4(2024)]{websight}
HuggingFaceM4.
\newblock Websight dataset.
\newblock \url{https://huggingface.co/datasets/HuggingFaceM4/WebSight}, 2024.

\bibitem[Kantharaj et~al.(2022)Kantharaj, Leong, Lin, Masry, Thakkar, Hoque, and Joty]{chart2text}
S.~Kantharaj, R.~T. Leong, X.~Lin, A.~Masry, M.~Thakkar, E.~Hoque, and S.~Joty.
\newblock Chart-to-text: A large-scale benchmark for chart summarization.
\newblock In S.~Muresan, P.~Nakov, and A.~Villavicencio, editors, \emph{Proceedings of the 60th Annual Meeting of the Association for Computational Linguistics (Volume 1: Long Papers)}, pages 4005--4023, Dublin, Ireland, May 2022. Association for Computational Linguistics.
\newblock \doi{10.18653/v1/2022.acl-long.277}.
\newblock URL \url{https://aclanthology.org/2022.acl-long.277}.

\bibitem[Kirillov et~al.(2023)Kirillov, Mintun, Ravi, Mao, Rolland, Gustafson, Xiao, Whitehead, Berg, Lo, et~al.]{sam}
A.~Kirillov, E.~Mintun, N.~Ravi, H.~Mao, C.~Rolland, L.~Gustafson, T.~Xiao, S.~Whitehead, A.~C. Berg, W.-Y. Lo, et~al.
\newblock Segment anything.
\newblock \emph{arXiv preprint arXiv:2304.02643}, 2023.

\bibitem[Kocetkov et~al.(2023)Kocetkov, Li, Allal, Li, Mou, Ferrandis, Jernite, Mitchell, Hughes, Wolf, Bahdanau, von Werra, and de~Vries]{DenisKocetkov2023}
D.~Kocetkov, R.~Li, L.~B. Allal, J.~Li, C.~Mou, C.~M. Ferrandis, Y.~Jernite, M.~Mitchell, S.~Hughes, T.~Wolf, D.~Bahdanau, L.~von Werra, and H.~de~Vries.
\newblock The stack: 3 tb of permissively licensed source code.
\newblock In \emph{Transactions on Machine Learning Research}, 2023.

\bibitem[Korthikanti et~al.(2023)Korthikanti, Casper, Lym, McAfee, Andersch, Shoeybi, and Catanzaro]{megatron3}
V.~A. Korthikanti, J.~Casper, S.~Lym, L.~McAfee, M.~Andersch, M.~Shoeybi, and B.~Catanzaro.
\newblock Reducing activation recomputation in large transformer models.
\newblock \emph{Proceedings of Machine Learning and Systems}, 5, 2023.

\bibitem[Krylov et~al.(2021)Krylov, Nosov, and Sovrasov]{krylov2021open}
I.~Krylov, S.~Nosov, and V.~Sovrasov.
\newblock Open images v5 text annotation and yet another mask text spotter.
\newblock In \emph{Asian Conference on Machine Learning}, pages 379--389. PMLR, 2021.

\bibitem[Kulkarni and Truelsen()]{wkhtmltopdf}
A.~Kulkarni and J.~Truelsen.
\newblock {wkhtmltopdf}.
\newblock \url{https://wkhtmltopdf.org/}.
\newblock Project maintained by Ashish Kulkarni, originally created by Jakob Truelsen. Accessed: 2024-02-22.

\bibitem[{LAION}(2023)]{laion-gpt4v}
{LAION}.
\newblock Gpt-4v dataset.
\newblock \url{https://huggingface.co/datasets/laion/gpt4v-dataset}, 2023.

\bibitem[Li et~al.(2023{\natexlab{a}})Li, Wang, Wang, Ge, Ge, and Shan]{li2023seed}
B.~Li, R.~Wang, G.~Wang, Y.~Ge, Y.~Ge, and Y.~Shan.
\newblock Seed-bench: Benchmarking multimodal llms with generative comprehension.
\newblock \emph{arXiv preprint arXiv:2307.16125}, 2023{\natexlab{a}}.

\bibitem[Li and Tajbakhsh(2023)]{scigraphqa-295k}
S.~Li and N.~Tajbakhsh.
\newblock Scigraphqa: A large-scale synthetic multi-turn question-answering dataset for scientific graphs, 2023.

\bibitem[Li et~al.(2020)Li, Li, He, Zheng, Li, and Guan]{widget-captioning}
Y.~Li, G.~Li, L.~He, J.~Zheng, H.~Li, and Z.~Guan.
\newblock Widget captioning: Generating natural language description for mobile user interface elements.
\newblock \emph{arXiv preprint arXiv:2010.04295}, 2020.

\bibitem[Li et~al.(2022)Li, Mao, Girshick, and He]{vitdet}
Y.~Li, H.~Mao, R.~Girshick, and K.~He.
\newblock Exploring plain vision transformer backbones for object detection.
\newblock In \emph{European Conference on Computer Vision}, pages 280--296. Springer, 2022.

\bibitem[Li et~al.(2023{\natexlab{b}})Li, Du, Zhou, Wang, Zhao, and Wen]{li2023evaluating}
Y.~Li, Y.~Du, K.~Zhou, J.~Wang, W.~X. Zhao, and J.-R. Wen.
\newblock Evaluating object hallucination in large vision-language models.
\newblock \emph{arXiv preprint arXiv:2305.10355}, 2023{\natexlab{b}}.

\bibitem[Lin et~al.(2023{\natexlab{a}})Lin, Yin, Ping, Lu, Molchanov, Tao, Mao, Kautz, Shoeybi, and Han]{lin2023vila}
J.~Lin, H.~Yin, W.~Ping, Y.~Lu, P.~Molchanov, A.~Tao, H.~Mao, J.~Kautz, M.~Shoeybi, and S.~Han.
\newblock Vila: On pre-training for visual language models.
\newblock \emph{arXiv preprint arXiv:2312.07533}, 2023{\natexlab{a}}.

\bibitem[Lin et~al.(2023{\natexlab{b}})Lin, Liu, Zhang, Gao, Qiu, Xiao, Qiu, Lin, Shao, Chen, et~al.]{lin2023sphinx}
Z.~Lin, C.~Liu, R.~Zhang, P.~Gao, L.~Qiu, H.~Xiao, H.~Qiu, C.~Lin, W.~Shao, K.~Chen, et~al.
\newblock Sphinx: The joint mixing of weights, tasks, and visual embeddings for multi-modal large language models.
\newblock \emph{arXiv preprint arXiv:2311.07575}, 2023{\natexlab{b}}.

\bibitem[Liu et~al.(2022{\natexlab{a}})Liu, Piccinno, Krichene, Pang, Lee, Joshi, Altun, Collier, and Eisenschlos]{liu2022matcha}
F.~Liu, F.~Piccinno, S.~Krichene, C.~Pang, K.~Lee, M.~Joshi, Y.~Altun, N.~Collier, and J.~M. Eisenschlos.
\newblock Matcha: Enhancing visual language pretraining with math reasoning and chart derendering.
\newblock \emph{arXiv preprint arXiv:2212.09662}, 2022{\natexlab{a}}.

\bibitem[Liu et~al.(2024{\natexlab{a}})Liu, Li, Li, Li, Zhang, Shen, and Lee]{llava-v1-6}
H.~Liu, C.~Li, Y.~Li, B.~Li, Y.~Zhang, S.~Shen, and Y.~J. Lee.
\newblock Llava-next: Improved reasoning, ocr, and world knowledge, January 2024{\natexlab{a}}.
\newblock URL \url{https://llava-vl.github.io/blog/2024-01-30-llava-next/}.

\bibitem[Liu et~al.(2024{\natexlab{b}})Liu, Li, Wu, and Lee]{llava-v1}
H.~Liu, C.~Li, Q.~Wu, and Y.~J. Lee.
\newblock Visual instruction tuning.
\newblock \emph{Advances in neural information processing systems}, 36, 2024{\natexlab{b}}.

\bibitem[Liu et~al.(2022{\natexlab{b}})Liu, Zhu, Zhu, Song, Ge, Chen, Qiao, Peng, Wu, and Wang]{liu2022taisu}
Y.~Liu, G.~Zhu, B.~Zhu, Q.~Song, G.~Ge, H.~Chen, G.~Qiao, R.~Peng, L.~Wu, and J.~Wang.
\newblock Taisu: A 166m large-scale high-quality dataset for chinese vision-language pre-training.
\newblock In S.~Koyejo, S.~Mohamed, A.~Agarwal, D.~Belgrave, K.~Cho, and A.~Oh, editors, \emph{Advances in Neural Information Processing Systems}, volume~35, pages 16705--16717. Curran Associates, Inc., 2022{\natexlab{b}}.
\newblock URL \url{https://proceedings.neurips.cc/paper_files/paper/2022/file/6a386d703b50f1cf1f61ab02a15967bb-Paper-Datasets_and_Benchmarks.pdf}.

\bibitem[Liu et~al.(2023{\natexlab{a}})Liu, Duan, Zhang, Li, Zhang, Zhao, Yuan, Wang, He, Liu, et~al.]{liu2023mmbench}
Y.~Liu, H.~Duan, Y.~Zhang, B.~Li, S.~Zhang, W.~Zhao, Y.~Yuan, J.~Wang, C.~He, Z.~Liu, et~al.
\newblock Mmbench: Is your multi-modal model an all-around player?
\newblock \emph{arXiv preprint arXiv:2307.06281}, 2023{\natexlab{a}}.

\bibitem[Liu et~al.(2023{\natexlab{b}})Liu, Li, Li, Yu, Huang, Peng, Liu, Chen, Li, Jin, et~al.]{liu2023hidden}
Y.~Liu, Z.~Li, H.~Li, W.~Yu, M.~Huang, D.~Peng, M.~Liu, M.~Chen, C.~Li, L.~Jin, et~al.
\newblock On the hidden mystery of ocr in large multimodal models.
\newblock \emph{arXiv preprint arXiv:2305.07895}, 2023{\natexlab{b}}.

\bibitem[Long et~al.(2022)Long, Qin, Panteleev, Bissacco, Fujii, and Raptis]{long2022towards}
S.~Long, S.~Qin, D.~Panteleev, A.~Bissacco, Y.~Fujii, and M.~Raptis.
\newblock Towards end-to-end unified scene text detection and layout analysis.
\newblock In \emph{Proceedings of the IEEE/CVF Conference on Computer Vision and Pattern Recognition}, 2022.

\bibitem[Lu et~al.(2021)Lu, Qiu, Chen, Xia, Zhao, Zhang, Yu, Liang, and Zhu]{lu2021iconqa}
P.~Lu, L.~Qiu, J.~Chen, T.~Xia, Y.~Zhao, W.~Zhang, Z.~Yu, X.~Liang, and S.-C. Zhu.
\newblock Iconqa: A new benchmark for abstract diagram understanding and visual language reasoning.
\newblock \emph{arXiv preprint arXiv:2110.13214}, 2021.

\bibitem[Lu et~al.(2022{\natexlab{a}})Lu, Mishra, Xia, Qiu, Chang, Zhu, Tafjord, Clark, and Kalyan]{lu2022learn}
P.~Lu, S.~Mishra, T.~Xia, L.~Qiu, K.-W. Chang, S.-C. Zhu, O.~Tafjord, P.~Clark, and A.~Kalyan.
\newblock Learn to explain: Multimodal reasoning via thought chains for science question answering.
\newblock In \emph{The 36th Conference on Neural Information Processing Systems (NeurIPS)}, 2022{\natexlab{a}}.

\bibitem[Lu et~al.(2022{\natexlab{b}})Lu, Mishra, Xia, Qiu, Chang, Zhu, Tafjord, Clark, and Kalyan]{scienceqa}
P.~Lu, S.~Mishra, T.~Xia, L.~Qiu, K.-W. Chang, S.-C. Zhu, O.~Tafjord, P.~Clark, and A.~Kalyan.
\newblock Learn to explain: Multimodal reasoning via thought chains for science question answering.
\newblock \emph{Advances in Neural Information Processing Systems}, 35:\penalty0 2507--2521, 2022{\natexlab{b}}.

\bibitem[Lu et~al.(2023)Lu, Bansal, Xia, Liu, Li, Hajishirzi, Cheng, Chang, Galley, and Gao]{lu2023mathvista}
P.~Lu, H.~Bansal, T.~Xia, J.~Liu, C.~Li, H.~Hajishirzi, H.~Cheng, K.-W. Chang, M.~Galley, and J.~Gao.
\newblock Mathvista: Evaluating mathematical reasoning of foundation models in visual contexts.
\newblock \emph{arXiv preprint arXiv:2310.02255}, 2023.

\bibitem[Mao et~al.(2016)Mao, Huang, Toshev, Camburu, Yuille, and Murphy]{refexp}
J.~Mao, J.~Huang, A.~Toshev, O.~Camburu, A.~L. Yuille, and K.~Murphy.
\newblock Generation and comprehension of unambiguous object descriptions.
\newblock In \emph{Proceedings of the IEEE conference on computer vision and pattern recognition}, pages 11--20, 2016.

\bibitem[Masry et~al.(2023)Masry, Kavehzadeh, Do, Hoque, and Joty]{unichart}
A.~Masry, P.~Kavehzadeh, X.~L. Do, E.~Hoque, and S.~Joty.
\newblock Unichart: A universal vision-language pretrained model for chart comprehension and reasoning.
\newblock \emph{arXiv preprint arXiv:2305.14761}, 2023.

\bibitem[Narayanan et~al.(2021)Narayanan, Shoeybi, Casper, LeGresley, Patwary, Korthikanti, Vainbrand, Kashinkunti, Bernauer, Catanzaro, et~al.]{megatron2}
D.~Narayanan, M.~Shoeybi, J.~Casper, P.~LeGresley, M.~Patwary, V.~Korthikanti, D.~Vainbrand, P.~Kashinkunti, J.~Bernauer, B.~Catanzaro, et~al.
\newblock Efficient large-scale language model training on gpu clusters using megatron-lm.
\newblock In \emph{Proceedings of the International Conference for High Performance Computing, Networking, Storage and Analysis}, pages 1--15, 2021.

\bibitem[Nayef et~al.(2017)Nayef, Yin, Bizid, Choi, Feng, Karatzas, Luo, Pal, Rigaud, Chazalon, et~al.]{nayef2017icdar2017}
N.~Nayef, F.~Yin, I.~Bizid, H.~Choi, Y.~Feng, D.~Karatzas, Z.~Luo, U.~Pal, C.~Rigaud, J.~Chazalon, et~al.
\newblock Icdar2017 robust reading challenge on multi-lingual scene text detection and script identification-rrc-mlt.
\newblock In \emph{2017 14th IAPR international conference on document analysis and recognition (ICDAR)}, volume~1, pages 1454--1459. IEEE, 2017.

\bibitem[OpenAI(2022)]{chatgpt}
OpenAI.
\newblock Chatgpt: Optimizing language models for dialogue.
\newblock 2022.
\newblock URL \url{https://openai.com/blog/chatgpt}.

\bibitem[OpenAI(2023{\natexlab{a}})]{gpt4}
OpenAI.
\newblock {GPT-4} technical report.
\newblock \emph{arXiv}, 2023{\natexlab{a}}.

\bibitem[OpenAI(2023{\natexlab{b}})]{gpt4v}
R.~OpenAI.
\newblock Gpt-4v(ision) system card.
\newblock 2023{\natexlab{b}}.

\bibitem[Rodriguez et~al.(2023)Rodriguez, Vazquez, Laradji, Pedersoli, and Rodriguez]{paper2figure100k}
J.~A. Rodriguez, D.~Vazquez, I.~Laradji, M.~Pedersoli, and P.~Rodriguez.
\newblock Ocr-vqgan: Taming text-within-image generation.
\newblock In \emph{Proceedings of the IEEE/CVF Winter Conference on Applications of Computer Vision}, pages 3689--3698, 2023.

\bibitem[Schaeffer et~al.(2024)Schaeffer, Miranda, and Koyejo]{schaeffer2024emergent}
R.~Schaeffer, B.~Miranda, and S.~Koyejo.
\newblock Are emergent abilities of large language models a mirage?
\newblock \emph{Advances in Neural Information Processing Systems}, 36, 2024.

\bibitem[Shazeer(2020)]{shazeer2020glu}
N.~Shazeer.
\newblock Glu variants improve transformer.
\newblock \emph{arXiv preprint arXiv:2002.05202}, 2020.

\bibitem[Shi et~al.(2017)Shi, Yao, Liao, Yang, Xu, Cui, Belongie, Lu, and Bai]{shi2017icdar2017}
B.~Shi, C.~Yao, M.~Liao, M.~Yang, P.~Xu, L.~Cui, S.~Belongie, S.~Lu, and X.~Bai.
\newblock Icdar2017 competition on reading chinese text in the wild (rctw-17).
\newblock In \emph{2017 14th iapr international conference on document analysis and recognition (ICDAR)}, volume~1, pages 1429--1434. IEEE, 2017.

\bibitem[Shoeybi et~al.(2019)Shoeybi, Patwary, Puri, LeGresley, Casper, and Catanzaro]{megatron}
M.~Shoeybi, M.~Patwary, R.~Puri, P.~LeGresley, J.~Casper, and B.~Catanzaro.
\newblock Megatron-lm: Training multi-billion parameter language models using model parallelism.
\newblock \emph{arXiv preprint arXiv:1909.08053}, 2019.

\bibitem[Singh et~al.(2021)Singh, Pang, Toh, Huang, Galuba, and Hassner]{singh2021textocr}
A.~Singh, G.~Pang, M.~Toh, J.~Huang, W.~Galuba, and T.~Hassner.
\newblock Textocr: Towards large-scale end-to-end reasoning for arbitrary-shaped scene text.
\newblock In \emph{Proceedings of the IEEE/CVF conference on computer vision and pattern recognition}, pages 8802--8812, 2021.

\bibitem[Su et~al.(2024)Su, Ahmed, Lu, Pan, Bo, and Liu]{su2024roformer}
J.~Su, M.~Ahmed, Y.~Lu, S.~Pan, W.~Bo, and Y.~Liu.
\newblock Roformer: Enhanced transformer with rotary position embedding.
\newblock \emph{Neurocomputing}, 568:\penalty0 127063, 2024.

\bibitem[Sun et~al.(2023)Sun, Yu, Cui, Zhang, Zhang, Wang, Gao, Liu, Huang, and Wang]{emu}
Q.~Sun, Q.~Yu, Y.~Cui, F.~Zhang, X.~Zhang, Y.~Wang, H.~Gao, J.~Liu, T.~Huang, and X.~Wang.
\newblock Generative pretraining in multimodality.
\newblock \emph{arXiv preprint arXiv:2307.05222}, 2023.

\bibitem[Sun et~al.(2019)Sun, Ni, Chng, Liu, Luo, Ng, Han, Ding, Liu, Karatzas, et~al.]{sun2019icdar}
Y.~Sun, Z.~Ni, C.-K. Chng, Y.~Liu, C.~Luo, C.~C. Ng, J.~Han, E.~Ding, J.~Liu, D.~Karatzas, et~al.
\newblock Icdar 2019 competition on large-scale street view text with partial labeling-rrc-lsvt.
\newblock In \emph{2019 International Conference on Document Analysis and Recognition (ICDAR)}, pages 1557--1562. IEEE, 2019.

\bibitem[Team et~al.(2023)Team, Anil, Borgeaud, Wu, Alayrac, Yu, Soricut, Schalkwyk, Dai, Hauth, et~al.]{gemini}
G.~Team, R.~Anil, S.~Borgeaud, Y.~Wu, J.-B. Alayrac, J.~Yu, R.~Soricut, J.~Schalkwyk, A.~M. Dai, A.~Hauth, et~al.
\newblock Gemini: a family of highly capable multimodal models.
\newblock \emph{arXiv preprint arXiv:2312.11805}, 2023.

\bibitem[Tong et~al.(2024)Tong, Liu, Zhai, Ma, LeCun, and Xie]{concat-dino}
S.~Tong, Z.~Liu, Y.~Zhai, Y.~Ma, Y.~LeCun, and S.~Xie.
\newblock Eyes wide shut? exploring the visual shortcomings of multimodal llms.
\newblock \emph{arXiv preprint arXiv:2401.06209}, 2024.

\bibitem[Touvron et~al.(2023{\natexlab{a}})Touvron, Lavril, Izacard, Martinet, Lachaux, Lacroix, Rozi{\`e}re, Goyal, Hambro, Azhar, et~al.]{llama}
H.~Touvron, T.~Lavril, G.~Izacard, X.~Martinet, M.-A. Lachaux, T.~Lacroix, B.~Rozi{\`e}re, N.~Goyal, E.~Hambro, F.~Azhar, et~al.
\newblock {LLaMA}: Open and efficient foundation language models.
\newblock \emph{arXiv preprint arXiv:2302.13971}, 2023{\natexlab{a}}.

\bibitem[Touvron et~al.(2023{\natexlab{b}})Touvron, Martin, Stone, Albert, Almahairi, Babaei, Bashlykov, Batra, Bhargava, Bhosale, Bikel, Blecher, Canton{-}Ferrer, Chen, Cucurull, Esiobu, Fernandes, Fu, Fu, Fuller, Gao, Goswami, Goyal, Hartshorn, Hosseini, Hou, Inan, Kardas, Kerkez, Khabsa, Kloumann, Korenev, Koura, Lachaux, Lavril, Lee, Liskovich, Lu, Mao, Martinet, Mihaylov, Mishra, Molybog, Nie, Poulton, Reizenstein, Rungta, Saladi, Schelten, Silva, Smith, Subramanian, Tan, Tang, Taylor, Williams, Kuan, Xu, Yan, Zarov, Zhang, Fan, Kambadur, Narang, Rodriguez, Stojnic, Edunov, and Scialom]{llama2}
H.~Touvron, L.~Martin, K.~Stone, P.~Albert, A.~Almahairi, Y.~Babaei, N.~Bashlykov, S.~Batra, P.~Bhargava, S.~Bhosale, D.~Bikel, L.~Blecher, C.~Canton{-}Ferrer, M.~Chen, G.~Cucurull, D.~Esiobu, J.~Fernandes, J.~Fu, W.~Fu, B.~Fuller, C.~Gao, V.~Goswami, N.~Goyal, A.~Hartshorn, S.~Hosseini, R.~Hou, H.~Inan, M.~Kardas, V.~Kerkez, M.~Khabsa, I.~Kloumann, A.~Korenev, P.~S. Koura, M.~Lachaux, T.~Lavril, J.~Lee, D.~Liskovich, Y.~Lu, Y.~Mao, X.~Martinet, T.~Mihaylov, P.~Mishra, I.~Molybog, Y.~Nie, A.~Poulton, J.~Reizenstein, R.~Rungta, K.~Saladi, A.~Schelten, R.~Silva, E.~M. Smith, R.~Subramanian, X.~E. Tan, B.~Tang, R.~Taylor, A.~Williams, J.~X. Kuan, P.~Xu, Z.~Yan, I.~Zarov, Y.~Zhang, A.~Fan, M.~Kambadur, S.~Narang, A.~Rodriguez, R.~Stojnic, S.~Edunov, and T.~Scialom.
\newblock Llama 2: Open foundation and fine-tuned chat models.
\newblock \emph{CoRR}, abs/2307.09288, 2023{\natexlab{b}}.
\newblock \doi{10.48550/arXiv.2307.09288}.
\newblock URL \url{https://doi.org/10.48550/arXiv.2307.09288}.

\bibitem[Veit et~al.(2016)Veit, Matera, Neumann, Matas, and Belongie]{veit2016coco}
A.~Veit, T.~Matera, L.~Neumann, J.~Matas, and S.~Belongie.
\newblock Coco-text: Dataset and benchmark for text detection and recognition in natural images.
\newblock \emph{arXiv preprint arXiv:1601.07140}, 2016.

\bibitem[Wang et~al.(2021)Wang, Li, Zhou, Chen, Grossman, and Li]{screen2words}
B.~Wang, G.~Li, X.~Zhou, Z.~Chen, T.~Grossman, and Y.~Li.
\newblock Screen2words: Automatic mobile ui summarization with multimodal learning.
\newblock In \emph{The 34th Annual ACM Symposium on User Interface Software and Technology}, pages 498--510, 2021.

\bibitem[Wang et~al.(2023{\natexlab{a}})Wang, Meng, Weng, He, Wu, and Jiang]{lvis-instruct4v}
J.~Wang, L.~Meng, Z.~Weng, B.~He, Z.~Wu, and Y.-G. Jiang.
\newblock To see is to believe: Prompting gpt-4v for better visual instruction tuning.
\newblock \emph{arXiv preprint arXiv:2311.07574}, 2023{\natexlab{a}}.

\bibitem[Wang et~al.(2023{\natexlab{b}})Wang, Lv, Yu, Hong, Qi, Wang, Ji, Yang, Zhao, Song, et~al.]{cogvlm}
W.~Wang, Q.~Lv, W.~Yu, W.~Hong, J.~Qi, Y.~Wang, J.~Ji, Z.~Yang, L.~Zhao, X.~Song, et~al.
\newblock Cogvlm: Visual expert for pretrained language models.
\newblock \emph{arXiv preprint arXiv:2311.03079}, 2023{\natexlab{b}}.

\bibitem[Wei et~al.(2023)Wei, Kong, Chen, Zhao, Ge, Yang, Sun, Han, and Zhang]{vary}
H.~Wei, L.~Kong, J.~Chen, L.~Zhao, Z.~Ge, J.~Yang, J.~Sun, C.~Han, and X.~Zhang.
\newblock Vary: Scaling up the vision vocabulary for large vision-language models.
\newblock \emph{arXiv preprint arXiv:2312.06109}, 2023.

\bibitem[Yang et~al.(2021)Yang, Panagopoulou, Lyu, Zhang, Yatskar, and Callison-Burch]{wikihow}
Y.~Yang, A.~Panagopoulou, Q.~Lyu, L.~Zhang, M.~Yatskar, and C.~Callison-Burch.
\newblock Visual goal-step inference using wikihow.
\newblock \emph{arXiv preprint arXiv:2104.05845}, 2021.

\bibitem[Ye et~al.(2023)Ye, Hu, Xu, Ye, Yan, Xu, Li, Tian, Qian, Zhang, et~al.]{ureader}
J.~Ye, A.~Hu, H.~Xu, Q.~Ye, M.~Yan, G.~Xu, C.~Li, J.~Tian, Q.~Qian, J.~Zhang, et~al.
\newblock Ureader: Universal ocr-free visually-situated language understanding with multimodal large language model.
\newblock \emph{arXiv preprint arXiv:2310.05126}, 2023.

\bibitem[Yu et~al.(2023{\natexlab{a}})Yu, Sun, Zhang, Cui, Zhang, Cao, Wang, and Liu]{capsfus}
Q.~Yu, Q.~Sun, X.~Zhang, Y.~Cui, F.~Zhang, Y.~Cao, X.~Wang, and J.~Liu.
\newblock Capsfusion: Rethinking image-text data at scale.
\newblock \emph{arXiv preprint arXiv:2310.20550}, 2023{\natexlab{a}}.

\bibitem[Yu et~al.(2023{\natexlab{b}})Yu, Yang, Li, Wang, Lin, Liu, Wang, and Wang]{yu2023mm}
W.~Yu, Z.~Yang, L.~Li, J.~Wang, K.~Lin, Z.~Liu, X.~Wang, and L.~Wang.
\newblock Mm-vet: Evaluating large multimodal models for integrated capabilities.
\newblock \emph{arXiv preprint arXiv:2308.02490}, 2023{\natexlab{b}}.

\bibitem[Yue et~al.(2023)Yue, Ni, Zhang, Zheng, Liu, Zhang, Stevens, Jiang, Ren, Sun, et~al.]{yue2023mmmu}
X.~Yue, Y.~Ni, K.~Zhang, T.~Zheng, R.~Liu, G.~Zhang, S.~Stevens, D.~Jiang, W.~Ren, Y.~Sun, et~al.
\newblock Mmmu: A massive multi-discipline multimodal understanding and reasoning benchmark for expert agi.
\newblock \emph{arXiv preprint arXiv:2311.16502}, 2023.

\bibitem[Zellers et~al.(2019)Zellers, Holtzman, Bisk, Farhadi, and Choi]{hellaswag}
R.~Zellers, A.~Holtzman, Y.~Bisk, A.~Farhadi, and Y.~Choi.
\newblock {HellaSwag}: Can a machine really finish your sentence?
\newblock In A.~Korhonen, D.~R. Traum, and L.~M{\`{a}}rquez, editors, \emph{Proceedings of the 57th Conference of the Association for Computational Linguistics, {ACL} 2019, Florence, Italy, July 28- August 2, 2019, Volume 1: Long Papers}, pages 4791--4800. Association for Computational Linguistics, 2019.
\newblock \doi{10.18653/v1/p19-1472}.
\newblock URL \url{https://doi.org/10.18653/v1/p19-1472}.

\bibitem[Zhang and Sennrich(2019)]{zhang2019root}
B.~Zhang and R.~Sennrich.
\newblock Root mean square layer normalization.
\newblock \emph{Advances in Neural Information Processing Systems}, 32, 2019.

\bibitem[Zhang et~al.(2024)Zhang, Du, Chen, Liang, Luo, Zheng, Zhu, Cheng, Xu, Guo, et~al.]{zhang2024cmmmu}
G.~Zhang, X.~Du, B.~Chen, Y.~Liang, T.~Luo, T.~Zheng, K.~Zhu, Y.~Cheng, C.~Xu, S.~Guo, et~al.
\newblock Cmmmu: A chinese massive multi-discipline multimodal understanding benchmark.
\newblock \emph{arXiv preprint arXiv:2401.11944}, 2024.

\bibitem[Zhang et~al.(2019)Zhang, Zhou, Jiang, Song, Li, Zhou, Wang, Wang, Liao, Yang, et~al.]{zhang2019icdar}
R.~Zhang, Y.~Zhou, Q.~Jiang, Q.~Song, N.~Li, K.~Zhou, L.~Wang, D.~Wang, M.~Liao, M.~Yang, et~al.
\newblock Icdar 2019 robust reading challenge on reading chinese text on signboard.
\newblock In \emph{2019 international conference on document analysis and recognition (ICDAR)}, pages 1577--1581. IEEE, 2019.

\bibitem[Zhang et~al.(2017)Zhang, Gueguen, Zharkov, Zhang, Seifert, and Kadlec]{UberText}
Y.~Zhang, L.~Gueguen, I.~Zharkov, P.~Zhang, K.~Seifert, and B.~Kadlec.
\newblock Uber-text: A large-scale dataset for optical character recognition from street-level imagery.
\newblock In \emph{SUNw: Scene Understanding Workshop - CVPR 2017}, Hawaii, U.S.A., 2017.
\newblock URL \url{http://sunw.csail.mit.edu/abstract/uberText.pdf}.

\bibitem[Zheng et~al.(2024)Zheng, Chiang, Sheng, Zhuang, Wu, Zhuang, Lin, Li, Li, Xing, et~al.]{zheng2024judging}
L.~Zheng, W.-L. Chiang, Y.~Sheng, S.~Zhuang, Z.~Wu, Y.~Zhuang, Z.~Lin, Z.~Li, D.~Li, E.~Xing, et~al.
\newblock Judging llm-as-a-judge with mt-bench and chatbot arena.
\newblock \emph{Advances in Neural Information Processing Systems}, 36, 2024.

\bibitem[Zhong et~al.(2023)Zhong, Cui, Guo, Liang, Lu, Wang, Saied, Chen, and Duan]{agieval}
W.~Zhong, R.~Cui, Y.~Guo, Y.~Liang, S.~Lu, Y.~Wang, A.~Saied, W.~Chen, and N.~Duan.
\newblock {AGIEval}: {A} human-centric benchmark for evaluating foundation models.
\newblock \emph{CoRR}, abs/2304.06364, 2023.
\newblock \doi{10.48550/arXiv.2304.06364}.
\newblock URL \url{https://doi.org/10.48550/arXiv.2304.06364}.

\bibitem[Zhu et~al.(2024)Zhu, Hessel, Awadalla, Gadre, Dodge, Fang, Yu, Schmidt, Wang, and Choi]{mmc4}
W.~Zhu, J.~Hessel, A.~Awadalla, S.~Y. Gadre, J.~Dodge, A.~Fang, Y.~Yu, L.~Schmidt, W.~Y. Wang, and Y.~Choi.
\newblock Multimodal c4: An open, billion-scale corpus of images interleaved with text.
\newblock \emph{Advances in Neural Information Processing Systems}, 36, 2024.

\end{thebibliography}

\newpage
\appendix

\section{Appendix}
\label{sec:appendix}

\begin{figure}[ht]
\centering
\includegraphics[width=1.0\textwidth]{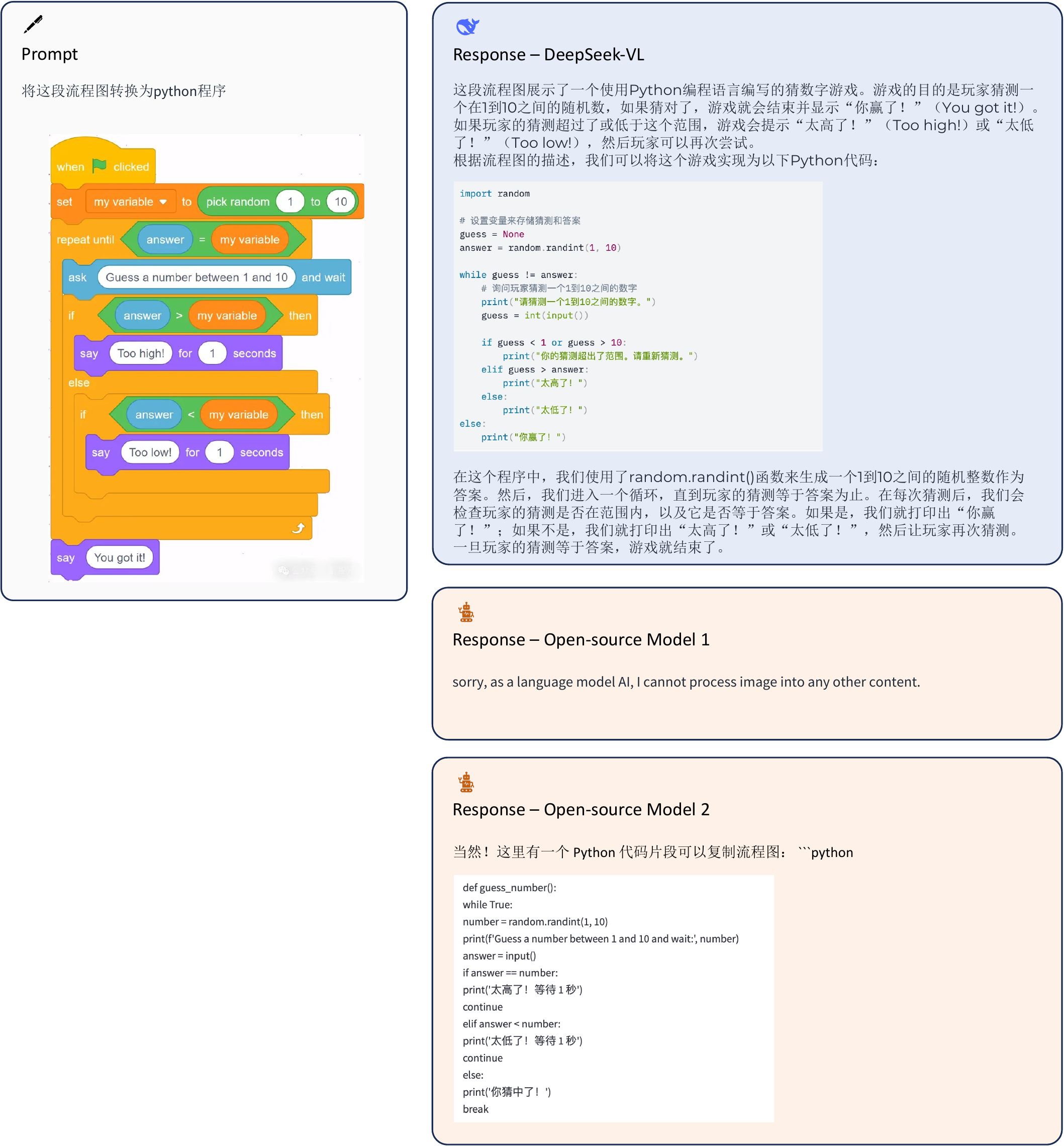}
\caption{Visualization results. DeepSeek-VL can understand children's programming diagrams from the real world and provide detailed and organized explanations.}
\label{fig:visualization_3}
\end{figure}

\begin{figure}[ht]
\centering
\includegraphics[width=0.85\textwidth]{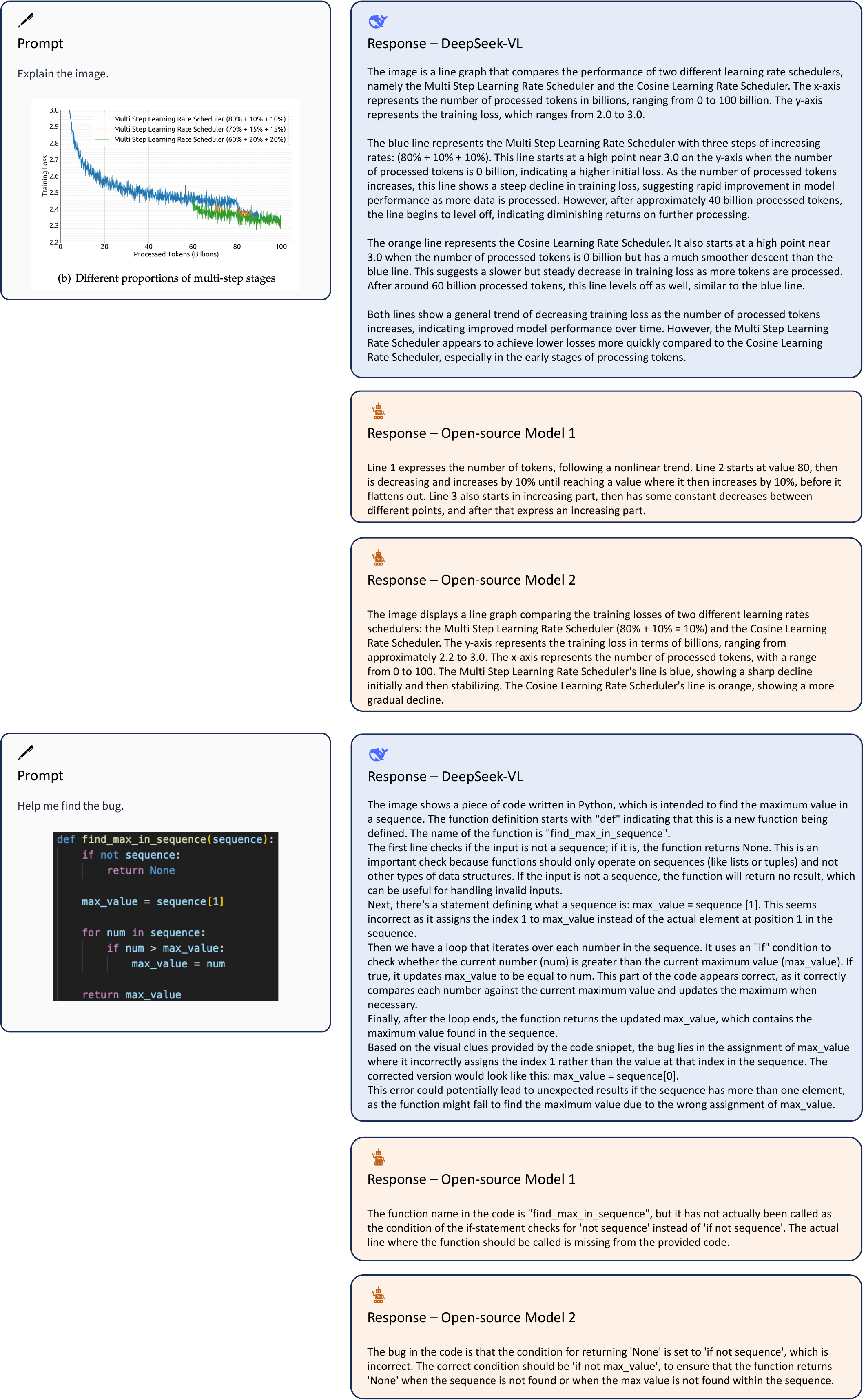}
\caption{Visualization results.  DeepSeek-VL has strong understanding capabilities for code and charts in the real world.}
\label{fig:visualization_1}
\end{figure}

\begin{figure}[ht]
\centering
\includegraphics[width=1.0\textwidth]{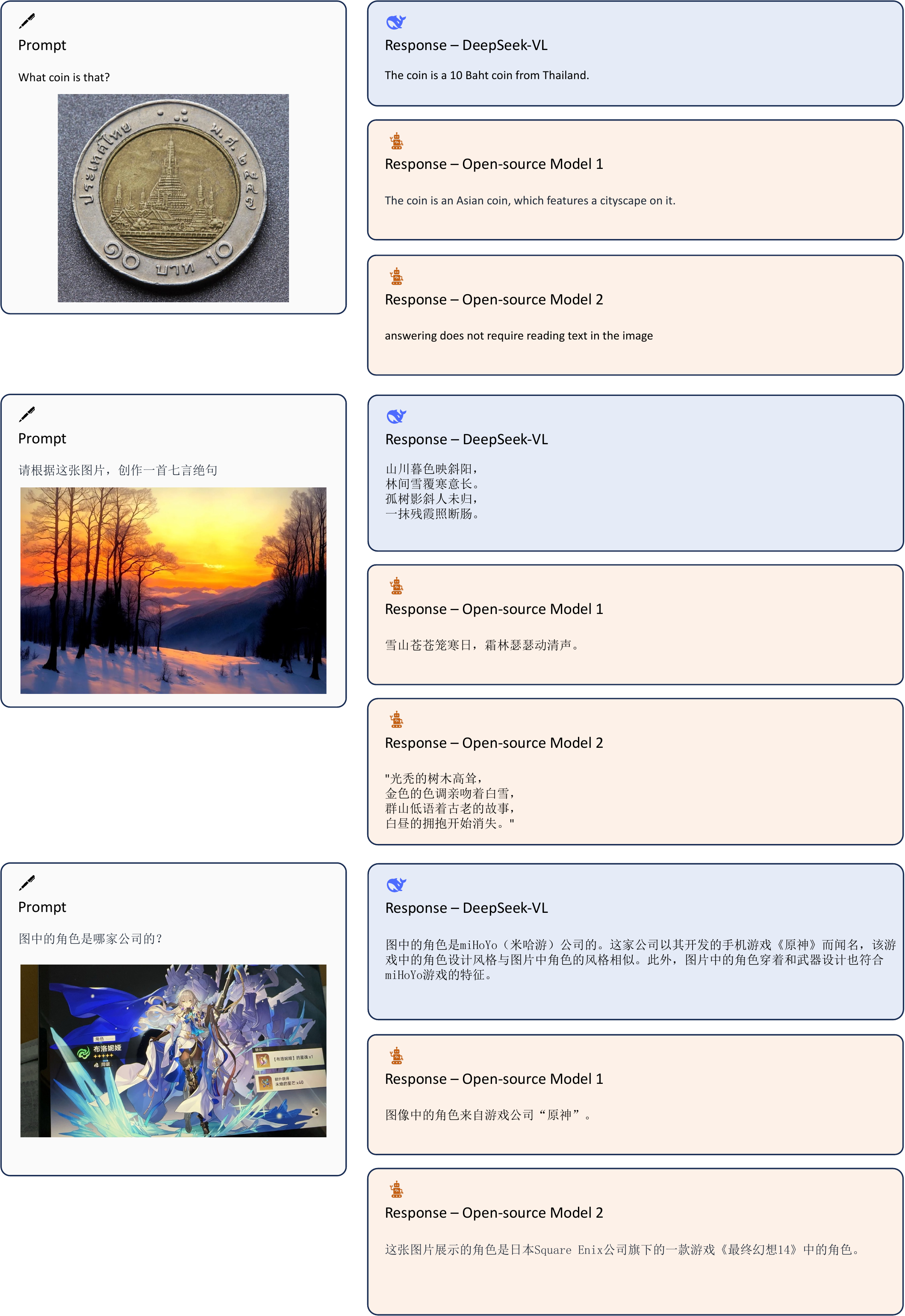}
\caption{Visualization results. DeepSeek-VL possesses extensive knowledge of the real world.}
\label{fig:visualization_2}
\end{figure}

\begin{figure}[ht]
\centering
\includegraphics[width=1.0\textwidth]{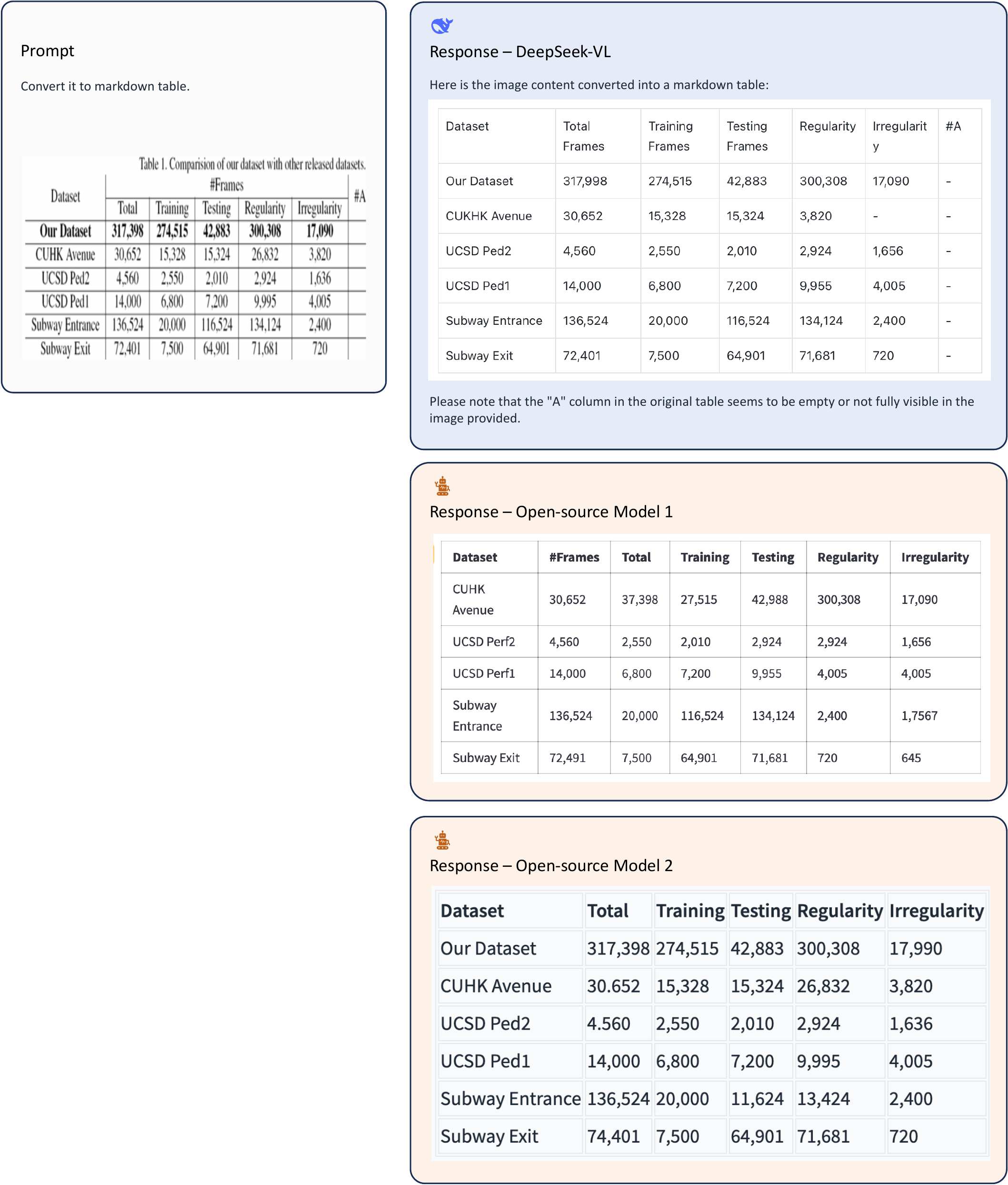}
\caption{Visualization results. 
DeepSeek-VL is capable of accurately reading the contents of real-world tables.}
\label{fig:visualization_4}
\end{figure}

\setcounter{figure}{0}
\makeatletter 
\renewcommand{\thefigure}{A\@arabic\c@figure}
\makeatother

\setcounter{table}{0}
\makeatletter 
\renewcommand{\thetable}{A\@arabic\c@table}
\makeatother

\end{CJK*}
\end{document}